\documentclass[11pt]{article}
\pdfoutput=1
\usepackage[
  margin=0.83in,
  headheight=12pt,
  headsep=25pt,
  includefoot,
  footskip=30pt,
]{geometry}
\usepackage{tikz}
\usepackage{graphicx}
\usepackage{comment}
\usepackage{listings}
\usepackage{xcolor}
\usepackage{inconsolata}
\usepackage{makecell}
\usepackage{tabularx} 
\usepackage{adjustbox}
\usepackage{rotating}

\definecolor{commentcolour}{rgb}{0.3,0.7,0.2}
\definecolor{backcolour}{rgb}{0.98,0.98,0.98}

\lstdefinelanguage{markdown}{
    comment=[l]{\#},
    morestring=[s]{```}{```},
    commentstyle=\color{commentcolour}\bfseries,
    stringstyle=\color{blue},
    basicstyle=\scriptsize\ttfamily,
    showstringspaces=false,
    breaklines=true,
    breakautoindent=false,
    breakindent=0pt,
    backgroundcolor=\color{backcolour},
}
\lstdefinestyle{mystyle}{
    morekeywords={self},
    basicstyle=\scriptsize\ttfamily,
    keywordstyle=\color{blue},
    commentstyle=\color{commentcolour}\bfseries,
    breaklines=true,
    breakautoindent=false,
    showstringspaces=false,
    backgroundcolor=\color{backcolour},
    stringstyle=\color{red},
}
\lstdefinelanguage{PythonPlus}[]{Python}{
  alsoother={@},
  morekeywords=[1]{,as,assert,nonlocal,with,yield,self,True,False,None} 
  morekeywords=[2]{,__init__,__add__,__mul__,__div__,__sub__,__call__,__getitem__,__setitem__,__eq__,__ne__,__nonzero__,__rmul__,__radd__,__repr__,__str__,__get__,__truediv__,__pow__,__name__,__future__,__all__,}, 
  morekeywords=[3]{,object,type,isinstance,copy,deepcopy,zip,enumerate,reversed,list,set,len,dict,tuple,range,xrange,append,execfile,real,imag,reduce,str,repr,}, 
  morekeywords=[4]{,Exception,NameError,IndexError,SyntaxError,TypeError,ValueError,OverflowError,ZeroDivisionError,}, 
  morekeywords=[5]{,ode,fsolve,sqrt,exp,sin,cos,arctan,arctan2,arccos,pi, array,norm,solve,dot,arange,isscalar,max,sum,flatten,shape,reshape,find,any,all,abs,plot,linspace,legend,quad,polyval,polyfit,hstack,concatenate,vstack,column_stack,empty,zeros,ones,rand,vander,grid,pcolor,eig,eigs,eigvals,svd,qr,tan,det,logspace,roll,min,mean,cumsum,cumprod,diff,vectorize,lstsq,cla,eye,xlabel,ylabel,squeeze,}, 
}

\usepackage{tikz}

\usepackage{amsmath} 
\usepackage{array} 
\usetikzlibrary{matrix, arrows}
\usepackage{amsmath,amssymb}
\usepackage{amsthm}
\usepackage{mathtools}
\usepackage{xspace}
\usepackage[noend]{algorithmic}
\usepackage[ruled,vlined]{algorithm2e}
\usepackage{url}
\usepackage{makeidx}
\usepackage{enumerate}
\usepackage{epstopdf}
\usepackage{booktabs}
\usepackage{color}
\usepackage[utf8]{inputenc}
\usepackage{thm-restate}
\usepackage{scalerel,stackengine}
\usepackage[shortlabels]{enumitem}
\usepackage{xr}
\usepackage{fancyvrb}
\usepackage{xcolor}
\usepackage{bold-extra}
\usepackage{arydshln}
\usepackage[small]{caption}
\usepackage{subcaption}
\usepackage[most]{tcolorbox}
\usepackage{fvextra}
\usepackage{float}
\usepackage{alltt}
\usepackage{soul}
\usepackage{MnSymbol,wasysym}
\usepackage{fancyvrb}
\usepackage{multirow}
\usepackage[bottom]{footmisc}
\usepackage[hidelinks]{hyperref}
\usepackage{diagbox}
\usepackage{mdframed}
\usepackage{todonotes}
\setuptodonotes{inline}
\usepackage{tikz}
\usetikzlibrary{shapes,calc,positioning}

\global

\tcbset{
  aibox/.style={
    width=\textwidth,
    top=0pt, bottom=0pt, left=5pt, right=5pt,
    colback=white,
    colframe=black,
    colbacktitle=black,
    enhanced,
    center,
    attach boxed title to top left={yshift=-0.1in,xshift=0.15in},
    boxed title style={boxrule=0pt,colframe=white,},
  }
}
\newtcolorbox{AIbox}[2][]{aibox,title=#2,#1}

\definecolor{aigold}{RGB}{244,210, 1} 
\definecolor{aigreen}{RGB}{210,244,211} 

\sethlcolor{aigreen}

\definecolor{aired}{RGB}{255,180,181}

\newcommand{\phivision}{Phi-3.5-Vision\xspace}

\newcommand{\datasetcell}[3]{\makecell{ \large #1  \\  \tiny (#2) \tiny #3   }  }

\newtcbox{\mybox}[1][green]{on line,
arc=0pt,outer arc=0pt,colback=#1!10!white,colframe=#1!50!black,
boxsep=0pt,left=0pt,right=0pt,top=0pt,bottom=0pt,
boxrule=0pt,bottomrule=0pt,toprule=0pt}

\newcommand{\phil}{\texttt{Phi-4-Mini}\xspace}
\newcommand{\phio}{\texttt{Phi-4-Multimodal}\xspace}

\begin{document}

\title{Phi-4-Mini Technical Report: Compact yet Powerful Multimodal Language Models via Mixture-of-LoRAs}

\author{Microsoft\footnote{hanyh@microsoft.com, youki@microsoft.com}}
\date{}

\maketitle

\begin{abstract}
We introduce \textbf{\phil} and \textbf{\phio}, compact yet highly capable language and multimodal models. \textbf{\phil} is a 3.8-billion-parameter language model trained on high-quality web and synthetic data, significantly outperforming recent open-source models of similar size and matching the performance of models twice its size on math and coding tasks requiring complex reasoning. This achievement is driven by a carefully curated synthetic data recipe emphasizing high-quality math and coding datasets. Compared to its predecessor, \texttt{Phi-3.5-Mini}, \textbf{\phil} features an expanded vocabulary size of 200K tokens to better support multilingual applications, as well as group query attention for more efficient long-sequence generation.
\textbf{\phio} is a multimodal model that integrates text, vision, and speech/audio input modalities into a single model. 
Its novel modality extension approach leverages LoRA adapters and modality-specific routers to allow multiple inference modes combining various modalities without interference. For example, it now ranks first in the OpenASR leaderboard to date, although the LoRA component of the speech/audio modality has just 460 million parameters. \textbf{\phio} supports scenarios involving (vision + language), (vision + speech), and (speech/audio) inputs, outperforming larger vision-language and speech-language models on a wide range of tasks.
Additionally, we experiment to further train \phil to enhance its reasoning capabilities.\footnote{Please note that reasoning-enhanced \phil is a separate model and currently in a preview stage and will not be released concurrently with \phil and \phio.}. Despite its compact 3.8-billion-parameter size, this experimental version achieves reasoning performance on par with or surpassing significantly larger models, including \texttt{DeepSeek-R1-Distill-Qwen-7B} and \texttt{DeepSeek-R1-Distill-Llama-8B}.

\end{abstract}

\section{Introduction}
The Phi family of models  \cite{abdin2024phi,abdin2024phi4} have shown that carefully curated and synthesized data enables Small Language Models (SLMs) to achieve highly competitive performance despite having a significantly smaller number of parameters. These models demonstrate comparable results to much larger models.
Building on the success of the Phi family of  language models, we  extend their capabilities to handle additional modalities — such as vision and audio, achieving significant progress akin to private models like GPT~\cite{hurst2024gpt}, Claude~\cite{anthropic2024claude}, and Gemini \cite{team2024gemini}.

In this report, we introduce \phio, a unified multimodal SLM that supports multiple inference modes combining various modalities (e.g., text-only, text + image, speech/audio, speech + image) within a single model checkpoint. \phio employs a novel ``mixture of LoRAs" technique, enabling multimodal capabilities by integrating modality-specific LoRAs while keeping the base language model entirely frozen. Our findings show this technique outperforms existing approaches (e.g., cross-attention designs \cite{alayrac2022flamingo,llama3}) and achieves comparable performance to fully fine-tuned models on multimodal benchmarks. Additionally, the design of \phio is highly extensible, allowing seamless integration of new LoRAs to support additional modalities without impacting existing ones.

Our training process comprises multiple stages, including language training (encompassing both pre-training and post-training) and then expansion of the language backbone to vision and speech/audio modalities. For the language model, we train \phil using high-quality, reasoning-rich text data. Notably, we include curated, high-quality code datasets to enhance performance on coding tasks. Once the language model training is complete, we freeze the language model and implement our ``Mixture of LoRAs'' technique to proceed with the multimodal training stage. Specifically, we train two additional LoRA modules alongside modality-specific encoders and projectors to enable vision-related tasks (e.g., vision-language and vision-speech) and speech/audio-related tasks (e.g., speech-language). Both of them contain pretraining and post-training stages for modality alignment and instruction finetuning, respectively.

We also explore the reasoning potential of \phil to create a compact yet powerful model that rivals substantially larger state-of-the-art reasoning systems, such as \texttt{DeepSeek-R1-Distill-Qwen-7B} and \texttt{DeepSeek-R1-Distill-Llama-8B} \cite{deepseekr1}. \\

The key contributions of this model are listed below.
\begin{enumerate}
    \item \textbf{Unified Multi-Modality Support}: In contrast to existing methods \cite{qwen2.5-VL, chen2024internvl} that employ separate models for different modalities, \phio is designed as a unified model capable of efficiently handling multiple modality scenarios. By leveraging the Mixture of LoRAs \cite{hu2022lora}, \phio extends multimodal capabilities while minimizing interference between modalities. This approach enables seamless integration and ensures consistent performance across tasks involving text, images, and speech/audio.
    \item \textbf{Remarkable Language Performance for the size}: The language models achieve state-of-the-art performance in natural language understanding and generation for its size category. It demonstrates exceptional reasoning and mathematical capabilities, making it well-suited for complex problem-solving and knowledge-based tasks.
    \item \textbf{Outstanding Code Understanding and Generation for the size}: The language models achieve state-of-the-art performance on code-related tasks within its size category. The model excels at tasks such as code synthesis, debugging, and documentation generation, empowering developers and aiding in software engineering workflows.
    \item \textbf{Superior Multi-Modal Capabilities for the size}: The model delivers state-of-the-art performance across multi-modal tasks for its size category, demonstrating robust integration of diverse data types. This includes tasks that involve combining images with text and speech modalities, enabling multi-modal reasoning.
    \item \textbf{Exceptional Speech and Audio Performance}: The model achieves strong performance especially on multilingual speech recognition and translation tasks, and is the first open-sourced model with speech summarization capability.
    \item  \textbf{Enhanced Reasoning Capabilities}: The reasoning-optimized version of \phil demonstrates superior reasoning abilities for a model in its size category.
\end{enumerate}

\section{Model architecture}

The \phil series comprises two state-of-the-art small models: a  language model (\phil) and a  multimodal model (\phio) that integrates language, vision, and speech/audio modalities.
All Phi-4-Mini models use the tokenizer \texttt{$\text{o200k}\_\text{base}$} tiktoken \footnote{\url{https://github.com/openai/tiktoken}} with a vocabulary size of 200,064 intended to support multilingual and multimodal input and output more efficiently. All models are based on decoder-only Transformer \cite{Vas17} and support 128K context length based on LongRoPE \cite{ding2024longrope}.

\subsection{Language model architecture}
\phil and \phio share the same language model backbone. \phil consist of 32 Transformer layers with hidden state size of 3,072 and tied input / output embedding which reduces the memory consumption significantly while providing much wider coverage of vocabularies compared \textbf{Phi-3.5}. Each Transformer block includes an attention mechanism based on Group Query Attention (GQA) \cite{ainslie2023gqa}, which optimizes key and value memory (KV cache) usage for long-context generation. Specifically, the model employs 24 query heads and 8 key/value heads, reducing KV cache consumption to one-third of its standard size. Additionally, in the RoPE configuration \cite{su2024roformer}, a fractional RoPE dimension is used, ensuring that 25\% of the attention head dimension remains position-agnostic. This design supports smoother handling of longer contexts. To determine the peak learning rate, we follow \cite{optimallr} with $LR^*(D)=BD^{-0.32}$ where $B$ is a constant we tune for this specific model and $D$ is the total number of training tokens. We fit $B$ by tuning across $D=12.5B,25B,37.5B,50B$.

\subsection{Multimodal model architecture}
\label{sec2.3-omni-arch}

To integrate vision as an input modality, numerous vision-language models have been developed, including the LLava series \cite{liu2024visual,liu2024llavanext,li2024llava}, QWenVL series \cite{bai2023qwen,wang2024qwen2}, InternVL series \cite{chen2024far,chen2024internvl,chen2024expanding}, InternLM-XComposer series \cite{zhang2023internlm,dong2024internlm}, Molmo \cite{deitke2024molmo}, and NVLM \cite{dai2024nvlm}. Similarly, for audio input, notable contributions include Qwen2-Audio \cite{chu2024qwen2}, InternLM-XComposer2.5-Omnilive \cite{zhang2024internlm}, InternOmni, Mini-Omni~\cite{xie2024mini}, and GLM4-Voice~\cite{zeng2024glm}.

However, in order to enable  modality-specific functionality, these multimodal models generally require fine-tuning the base language model, which often diminishes its original language capabilities. Consequently, supporting diverse input signals without compromising quality necessitates deploying multiple models—a particularly challenging limitation for resource-constrained devices. To address this, LLama-Vision \cite{dubey2024llama} adopts a strategy inspired by Flamingo~ \cite{alayrac2022flamingo}, adding extra cross-attention layers while preserving the core language model. However, this approach will result in reduced performance on vision-language benchmarks compared to fully fine-tuned models. To fill the performance gap, NVLM \cite{dai2024nvlm} further explores a hybrid framework, employing joint supervised fine-tuning with high-quality text SFT data. Yet, this approach only examines limited language benchmarks and does not address additional training stages often required after SFT.

\begin{figure}
    \centering
    \includegraphics[width=1.0\textwidth]{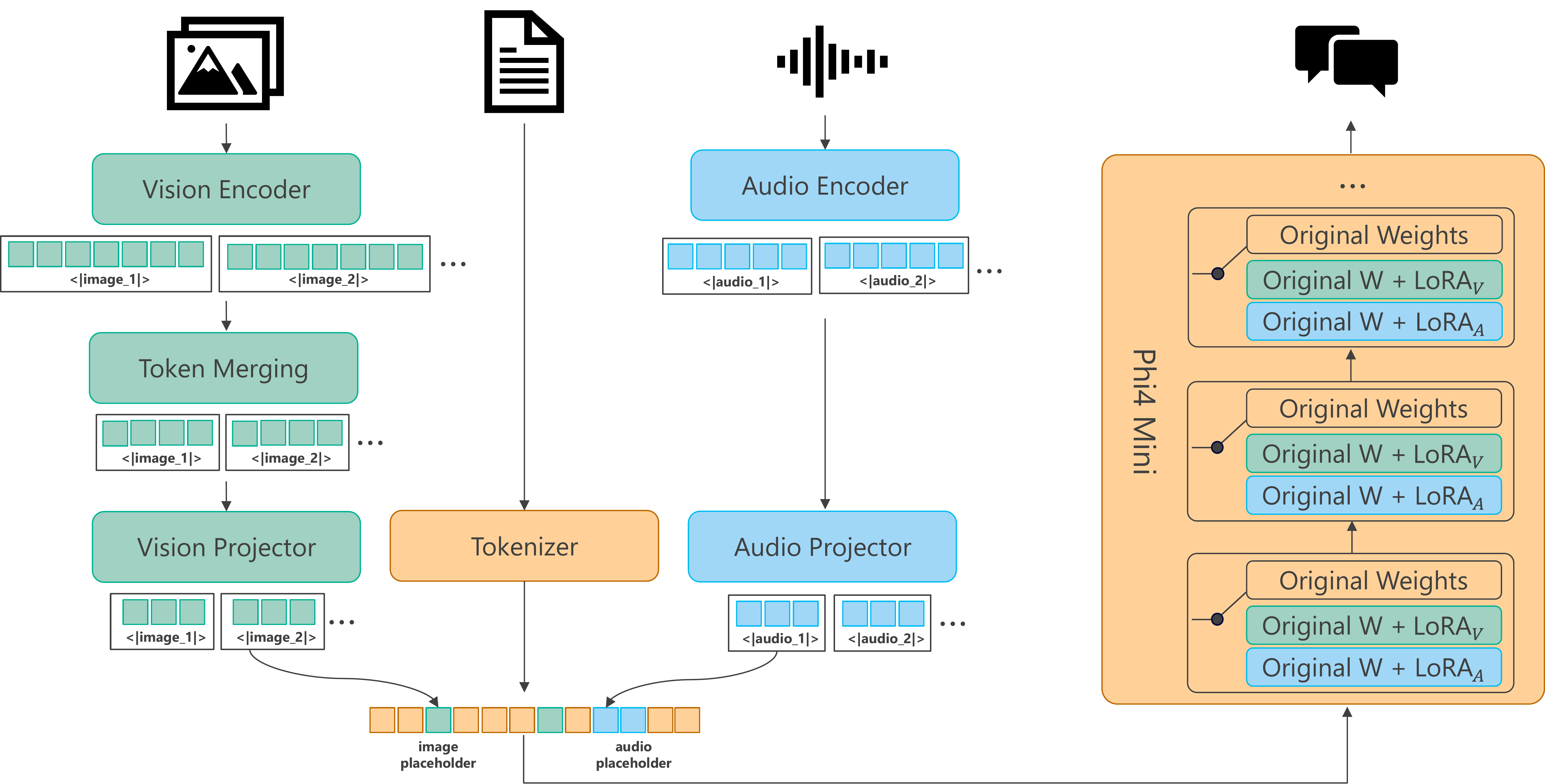}
    \caption{A overview of the Multimodal architecture for \phio}
    \label{fig:omni}
\end{figure}

We adopt the mixture of LoRAs design for our \phio architecture to support variant multi-modality use cases. Different LoRAs are trained to handle interactions between different modalities. Our \phio supports a vast range of tasks, including single/multiple images QA/summarization, video QA/summarization, vision-speech tasks, speech QA/summarization/translation/recognition, and audio understanding, while maintains the original language model performance.
\subsubsection{Modality Details}
\label{ssec:omni-model}

\paragraph{Vision modality.} The vision modality is implemented with an image encoder, a projector to align the vision and text embeddings and a LoRA adaptor. The vision encoder is based on SigLIP-400M that is finetuned with LLM2CLIP~\cite{huang2024llm2clip} on large scale image-text pairs with resolution $448\times448$ . The projector is a 2-layer MLP that maps the vision features dimension to the text embedding dimension. Extra LoRA is added on all the linear layers in the language decoder and only deployed in the supervised fine tuning (SFT) stage. The image encoder and projector introduce 440M model parameters while the vision adapter $LoRA_V$ consumes another 370M model parameters.

In order to enable the model to process images with diverse resolution effectively and efficiently, we proposed a new dynamic multi-crop strategy. Specifically, given a target image, we first compute the crop number for each side by dividing the original size by the crop-size, i.e. $\lceil \frac{H}{C}\rceil \times \lceil \frac{W}{C}\rceil$, where $H, W, C$ are the image height, width and crop size respectively.  If the total crop number is within the maximum number, i.e., $16$ in the pretraining stage and $36$ in SFT, we just slightly resize the image to let it fit the size given by the computed image crops. Otherwise, we will leverage the strategy proposed in InternVL2 \cite{chen2024internvl} that find the crop number by matching the best aspect ratio. Compared to InternVL2, the key benefits of our strategy is to avoid resizing one small image (e.g., $28\times 448$) to unreasonable large size when looking for the closest image aspect ratio.

\paragraph{Speech and Audio Modality:} The speech/audio inputs we used are 80-dim log-Mel filter-bank features with the frame rate of 10ms. To enable \phio speech and audio functions, we connect a pre-trained audio encoder and \phil through an audio adapter. In addition, LoRA is applied on the language decoder to improve the performance of speech/audio benchmarks while preserving the text capability. The introduced modules for the speech/audio modality include:
\begin{itemize}
    \item An audio encoder, which consists 3 convolutions layers and 24 conformer blocks~\cite{conformer20} with 1024 attention dimensions, 1536 feed-forward dimensions, and 16 attention heads. The convolution layers contribute to a sub-sampling rate of 8, and thus 80ms token rate for the language decoder. \item An audio projector, which is a 2-layer MLP that maps the 1024-dim speech features to the text embedding space of 3072 dimensions, similar to the vision projector. 
    \item $LoRA_A$ that has been applied to all attention and MLP layers in \phil with a rank of 320. 
\end{itemize}

The audio encoder and projector introduce 460M parameters while $LoRA_A$ consumes another 460M parameters. Note that the speech token rate is 80ms, indicating 750 tokens for 1-minute audio.

\subsubsection{Training Pipeline}
\label{ssec:training_details}
The multimodal training stages include vision training, speech/audio training and vision-speech joint training.

\paragraph{Vision Training.} The overall training pipeline for multimodal learning consists of vision training, speech and audio training, and joint vision-audio training. Vision training follows a four-stage process: 1) Projector Alignment stage: initially, only the projector is trained using caption data to align vision and text embeddings while preserving the pretrained representation of the vision encoder. 2) Joint Vision Training stage: Next, the projector and vision encoder are jointly trained on the full vision pretraining dataset to enhance key vision capabilities, such as OCR and dense understanding. 3) Generative Vision-Language Training stage: LoRA is then deployed on the language decoder and trained alongside the vision encoder and projector using curated single-frame SFT data, equipping the model with generative capabilities for vision-language inputs. 4) Multi-Frame Training stage: Finally, the model is trained on multi-frame SFT data with the vision encoder frozen, extending the context length coverage to 64k and enabling multi-image and temporal understanding.

\paragraph{Speech and Audio Training.} With the \phil language model, we conduct a two-stage paradigm for speech and audio training, also known as speech/audio pre-training and post-training. In the pre-training stage, we use large-scale automatic speech recognition (ASR) data to align the audio encoder and \phil in the semantic space. In this stage, the encoder and projector is updated with a learning rate of 4e-5 for 50k steps while the language decoder is frozen. We initialize the audio encoder with a pre-trained encoder from the attention-based encoder decoder (AED) ASR model.

After the pre-training stage, the model can only perform the ASR task. To unlock the instruction following capability of \phio for variety of speech and audio tasks, we continue to train the model with about 100M curated speech and audio SFT samples (after weighted up) as the speech post-training stage. Please refer to Section~\ref{sssec3.3:speech-post-training-data} for data details. In speech/audio post-training, the audio encoder is frozen. We update the audio projector and $LoRA_A$ with a learning rate of 1e-4 for another 50k steps. We consider different maximum audio lengths for different tasks in post-training. For speech summarization task, we train up to 30-minute audio (22.5k tokens). For other tasks, the maximum audio exposed in training is 30s (375 tokens). If we consider the 128k context length for language decoder, theoretically \phio can support a maximum 2.8 hours of audio as out of the box inference. It is worth noting that we have not fine tuned the model on such long audio data and it may need  further fine tuning to practically support such use cases.

\paragraph{Vision-speech Joint Training.}
The vision-speech joint training is conducted after vision post-training and speech post-training. We freeze the language base model, audio encoder, and audio projector, while finetuning the vision adapter $LoRA_V$, vision encoder, and the vision projector. In this stage, we train the model mainly on vision-speech SFT data but we also include a mixture of language and vision post-training data to maintain the corresponding performance. 

\paragraph{Reasoning Training}
Recent studies have suggested that training a robust reasoning model only requires a small amount of high-quality data, such as LIMO \cite{limo} and S1K \cite{s1}. However, we propose a fundamentally different training paradigm for SLM: we need to conduct a pre-training phase on extensive reasoning data to capture general reasoning chains, and then perform careful fine-tuning on curated SFT or preference data.
The continued training of \phil for reasoning proceeds in three distinct stages. 1) First, building on \phil, the model is pre-trained on approximately 60 billion reasoning CoT tokens generated by frontier reasoning  LLMs, after which rejection sampling is employed to filter out incorrect outputs. This allows the reasoning extension of \phil to learn the reasoning chains produced by these models. 2) In the second stage, the model is fine-tuned on a smaller but carefully curated dataset of around 200K high-quality CoT samples, chosen to cover diverse domains and varying difficulty levels. 3) Roll-Out DPO: finally, in the third stage, we label filtered incorrect outputs as “dis-preferred” and their corrected counterparts as `preferred', compiling a new dataset of 300K preference samples for DPO training.

\section{Data and training details}

\subsection{Language training data}
\subsubsection{Pre-training data}
Compared with Phi-3.5-Mini, we improved the quality of the pre-training data from several key aspects:
\begin{enumerate}
    \item \textit{Better data filtering}: By using an enhanced quality classifier, which is trained on a larger curated dataset consisting of cleaner positive and negative samples, we end up with better filtering quality across multiple languages with various aspects (e.g. toxic, obscure, scientific, etc.), leading to a more comprehensive and controllable filtering strategy overall.
    \item \textit{Better math and coding data}: For the math and coding data, we have augmented our original data with a specific instruction-based math and coding data set. This enhancement has resulted in effective results in math, coding and reasoning.
    \item \textit{Better synthetic data}: we incorporated Phi-4 synthetic data ~\cite{abdin2024phi4} into this model training with the same processing and decontamination.
    \item \textit{Better data mixture}: With the better classifiers, we re-tuned the data mixture with ablation experiments. Especially we increased the ratio for the reasoning data. That gives us a boost for the model quality.
\end{enumerate}

With these techniques, we built the 5 trillion pre-training data corpus, which is larger and in higher quality compared to the \textbf{Phi-3.5-Mini}. 

\subsubsection{Post-training data}
Compared to Phi-3.5-Mini, \phil includes a significantly larger and more diverse set of function calling and summarization data. Additionally, we synthesize a substantial amount of instruction-following data to enhance the model’s instruction-following capabilities. For coding, we incorporate extensive code completion data, including tasks that require the model to generate missing code in the middle of an existing code snippet. This challenges the model to understand both the requirements and the existing context, leading to significant performance improvements.

\subsubsection{Reasoning training data}
We generate a large volume of synthetic chain-of-thought (CoT) data from larger reasoning models, covering diverse domains and difficulty levels. During sampling, we employ both rule-based and model-based rejection methods to discard incorrect generations and feed them back for resampling. Also, we label correct sampled answers as preferred generations and incorrect ones as dis-preferred, and create the DPO data. This data has been utilized exclusively for the experimental reasoning model and has not been applied to the officially released checkpoint \phil.

\subsection{Vision-language training data}
The \phio model's pre-training phase involves a rich and varied dataset, encompassing interleaved image-text documents, image-text pairs, image grounding data, synthetic datasets from OCR of PDFs and realistic images, and synthesized datasets for chart comprehension. During this phase, the model's primary focus is on predicting the next token, concentrating solely on text tokens and disregarding any loss associated with image tokens. The pre-training process involves a total of 0.5T tokens, combining both visual and textual elements. Additionally, the maximum image resolution is capped at 1344x1344, as most training images are smaller than this size. For supervised fine-tuning (SFT), we utilized a combination of a text SFT dataset, publicly available multimodal instruction tuning datasets, and large-scale in-house multimodal instruction tuning datasets that we developed. These datasets span diverse domains and tasks, including general natural image understanding, chart, table, and diagram comprehension and reasoning, PowerPoint analysis, OCR, multi-image comparison, video summarization, and model safety. Collectively, the multimodal SFT data comprises approximately 0.3T tokens.

\subsection{Vision-speech training data}For vision-speech data, \phio model is trained on a diverse set of synthetic vision-speech data, covering single-frame and multi-frame scenarios. Specifically, we reuse a subset of vision-language SFT data and run in-house text-to-speech (TTS) engine to convert the user queries from texts to audios. This subset is carefully selected to avoid certain datasets where the queries are not suitable to read out in speech. We also measure the quality of the synthetic speech by transcribing the audio with in-house ASR model and calculating the word error rate (WER) between original text and transcription. Our final vision-speech data is generated with the WER-based filtering to ensure the quality.

\subsection{Speech and Audio Training Data}

The training data for speech/audio functions can be categorized into two types: 1) pre-training data with ASR transcriptions to provide a strong alignment between the speech and text modalities; 2) post-training data to unlock the instruction-following capability of \phio with the speech/audio modality involved. The post-training data covers a variety of tasks, including automatic speech recognition (ASR), automatic speech translation (AST), speech question answering (SQA), spoken query question answering (SQQA), speech summarization (SSUM), and audio understanding (AU).

\subsubsection{Pre-training Data}
\label{sssec3.3:speech-pretraining-data}
Despite that the audio encoder is initialized from a well-trained ASR model as mentioned in Sec.~\ref{sec2.3-omni-arch}, the speech and text latent spaces differ. To pre-train the adapter and reduce the modality gap between the speech and text sequences, we curate a dataset of approximately 2M hours of anonymized in-house speech-text pairs with strong/weak ASR supervisions, covering the eight supported languages~\footnote{The speech interface supports the following 8 languages: Chinese, English, French, German, Italian, Japanese, Portuguese, and Spanish.}. 

\subsubsection{Post-training Data}
\label{sssec3.3:speech-post-training-data}
Following language post-training paradigm, we curate SFT data for speech/audio post-training, aiming for unlocking the instruction-following capability with speech/audio as query or context. We use both the real and synthetic speech/audio data during speech post-training, covering the majority of speech and audio understanding tasks. All the SFT data are formatted as: \\

$<|user|><audio>\{\text{task prompt\}}<|end|><|assistant|>\{\text{label}\}<|end|>$ \\
where task prompt is to describe each task in the natural language description and it is null for the SQQA task.

\paragraph{Speech Recognition Data.} ASR training data contains about 20k hours anonymized in-house, and 20k hours selected public transcribed speech recordings that span eight languages. The weighted ASR training data contributes to 28M SFT examples.

\paragraph{Speech Translation Data.} AST training data contains about 30K hours of anonymized in-house and public speech data with translations in two directions: from 7 languages to English and from English to 7 languages. This data contains both supervised and synthetic translation from a machine translation model. The AST data is created with two formats: direct ST and ASR + translation in a Chain-of-thoughts (CoT) manner, contributing to 28M weighted training examples in post-training.

\paragraph{Speech and Spoken Query Question Answering Data.} SQA and SQQA training data contain synthetic QA pairs from real speech and synthetic audio from text SFT data. 
\begin{itemize}
    \item Synthetic QA pairs for SQA: To enable SQA capability, we reuse the speech-transcript pairs in the ASR training data and prompt the language model to generate multiple text QA pairs for each transcript. The low-quality QA pairs are filtered during training.
    \item Synthetic spoken query (audio) for SQQA: SQA is tasked to respond speech context plus text query. Responding to spoken query directly is also an important capability for \phio. Consequently, We sample from the language post-training data and convert the text query to audio query using our internal zero-shot TTS system.  
    \item Synthetic LM response for SQQA: Similar to~\cite{fathullah2024audiochatllama}, we synthetically generate responses for speech prompts by prompting the language model with the ASR transcripts of those prompts. The LM response data can improve the SQQA robustness of \phio in real scenarios because of more diverse spoken queries sampled from the ASR training data.
\end{itemize}
The total SQA and SQAQA data contribute to 26M weighted SFT examples.

\paragraph{Speech Summarization Data.} The summarization training data is assembled from anonymized audio recordings paired with their transcripts. The audio consists of multi-speaker conversational speech that spans a range of topics. Rather than dividing the audio into shorter segments, we maintain its full length up to a maximum of 30 minutes. To construct query-summary pairs for each audio clip, we use GPT-4 to generate a variety of queries and their respective summaries based on the transcripts. For each audio clip, the summarization queries address specific or general aspects of the conversation and vary in format, including length (number of words or sentences) and structure (summaries formatted as bullet points, JSON, or email). The weighted dataset contributes to 1M SFT examples with English speech only.

\paragraph{Audio Understanding Data.} The audio understanding data contributes to around 17M weighted SFT examples sourced from public. The dataset is created in the form of (audio, question, answer) tuples, where “audio” contains speech, audio, and music inputs. Similar to ~\cite{gong_ltuas}, the question and answer pairs are generated from GPT4 based on audio transcripts and/or meta information. \\

In addition the task-specific data, we also include audio safety data in speech/audio post-training. Please refer to Sec.~\ref{ssec:audio-safety} for the details of audio safety data. For all the public data, we utilize our Azure PII Detector\footnote{\url{https://learn.microsoft.com/en-us/azure/ai-services/language-service/personally-identifiable-information/overview}} to identify and handle Personally Identifiable Information (PII). The training examples with PII detected are removed for privacy concerns.

\section{Evaluation}

\subsection{Multimodal Benchmarks}

\begin{table}[ht!]
\begin{center}
\begin{adjustbox}{width=1.0\textwidth,center}
\begin{tabular}{ c||ccccccccccc } 
\hline
&\makecell{\phio\\ \footnotesize 5.6B } & \makecell{Phi-3.5-Vision\\ \footnotesize 4.2B } &  \makecell{Qwen2.5-VL-3B\\ \footnotesize 3.8B\\} & \makecell{InternVL2.5-4B\\ \footnotesize 3.7B} &
\makecell{Qwen2.5-VL-7B\\ \footnotesize 8.3B \\ } & \makecell{InternVL2.5-8B\\ \footnotesize 8.1B} &\makecell{Gemini-2.0-Flash-\\Lite-prv-02-05 }&\makecell{Gemini-2.0-\\Flash}  & \makecell{Claude-3.5\\-Sonnet}   &  \makecell{GPT-4o\\-mini} &  \makecell{GPT-4o\\ \footnotesize -}  \\

\hline & \\[-1.5ex]

\datasetcell{MMMU}{\scriptsize val}{\cite{yue2023mmmu}} & 55.1 & 43.0 & 47.0 & 48.3& 51.8 & 50.6  & 54.1 & 64.7& 55.8 & 52.1 & 61.7\\
\datasetcell{MMMUPro} {\scriptsize standard/vision}{\cite{yue2024mmmu}} & \makecell{38.5 \\ \footnotesize(39.7/37.3)} & \makecell{21.8 \\ \footnotesize(25.5/18.0)} & \makecell{29.9 \\ \footnotesize(31.8/28.0)} & \makecell{32.4 \\ \footnotesize(36.1/28.6)}& \makecell{38.7 \\ \footnotesize(39.5/37.9)}& \makecell{34.4 \\ \footnotesize(39.0/29.8)}  & \makecell{45.1 \\ \footnotesize(45.8/44.3)}&  \makecell{54.4 \\ \footnotesize(57.1/51.6)} & \makecell{54.3 \\ \footnotesize(56.5/52.1)} & \makecell{40.8 \\ \footnotesize(44.0/37.7)} & \makecell{53.0 \\ \footnotesize(55.3/50.7)}\\
\datasetcell{ScienceQA}{\scriptsize test}{\cite{lu2022learn}}  & 97.5& 91.3 & 79.4 & 96.2 & 87.7 & 97.3  & 85.0 & 88.3 & 81.2 & 84.0 & 88.2\\
\datasetcell{MathVista}{\scriptsize testmini}{\cite{lu2024mathvista}} & 62.4& 43.9& 60.8 & 51.2& 67.8 & 56.7 & 57.6 & 47.2 & 56.9  & 38.8 & 56.1\\
\datasetcell{Inter-GPS}{\scriptsize test}{\cite{lu2021intergps}} & 48.6 & 36.3 & 48.3& 53.7& 52.7 & 54.1 & 57.9 & 65.4 & 47.1 & 39.9 & 49.1\\
\datasetcell{MMBench}{\scriptsize dev-en}{\cite{liu2024mmbench}} & 86.7& 81.9 & 84.3 & 86.8 & 87.8 & 88.2  & 85.0 & 90.0 & 86.7 & 83.8 & 89.0\\
\datasetcell{POPE}{\scriptsize test}{ \cite{li2023evaluating}} & 85.6 & 86.1 & 87.9 & 89.4 & 87.5 & 89.1  & 87.5 & 88.0 & 82.6 & 83.6 & 86.5\\
\datasetcell{AI2D}{\scriptsize test}{\cite{kembhavi2016diagram}} & 82.3& 78.1& 78.4 & 80.0& 82.6& 83.0 & 77.6 & 82.1 & 70.6 & 75.2 & 83.8\\
\datasetcell{ChartQA}{\scriptsize test}{\cite{masry-etal-2022-chartqa}} & 81.4 & 81.8 & 80.0 & 79.1& 85.0 & 81.0 & 73.0 & 79.0 & 78.4 & 54.5 & 75.1 \\
\datasetcell{TextVQA}{\scriptsize test}{\cite{singh2019vqa}} & 75.6 & 72.0 & 76.8 & 70.9& 77.7& 74.8&72.9 & 74.4 & 58.6 & 70.9 & 73.1\\
\datasetcell{DocVQA}{\scriptsize test}{\cite{mathew2021docvqa}} & 93.2 & 69.3 & 93.9 & 91.6& 95.7 & 93.0 & 91.2 & 92.1 & 95.2 & 84.2 & 90.9\\
\datasetcell{InfoVQA}{\scriptsize test}{\cite{mathew2022infographicvqa}} & 72.7 & 36.6& 77.1 & 72.1 & 82.6 & 77.6  & 73.0 & 77.8 & 74.3 & 59.5 & 71.9 \\
{OCRBench}\\{\cite{liu2024ocrbench}} & 84.4 & 63.8 & 82.2 & 71.6 & 87.7 & 74.8 & 75.7 & 81.0 & 77.0 & 77.1 & 77.7\\

\hline & \\[-1.5ex]
\datasetcell{BLINK}{\scriptsize test}{\cite{fu2025blink}} & 61.3 & 57.0 & 48.1 & 51.2 & 55.3 & 52.5 & 59.3 & 64.0 & 56.9 & 51.9 & 62.4\\
\datasetcell{VideoMME-16Frame}{\scriptsize test}{\cite{fu2024video}} & 55.0& 50.8 & 56.5 & 57.3& 58.2 & 58.7 & 58.8& 65.5 & 60.2 & 61.2 & 68.2\\
\hline & \\[-1.5ex]
\textbf{Average} & \textbf{72.0} & \textbf{60.9} & \textbf{68.7} & \textbf{68.8} & \textbf{73.3} & \textbf{71.1} & \textbf{70.2} & \textbf{74.3} & \textbf{69.1} & \textbf{63.8} & \textbf{72.4}\\
\hline
\end{tabular}
\end{adjustbox}
\end{center}
\vspace{-1em}
\caption{Comparison results on public vision-language benchmarks. All the reported numbers are produced with the exact same internal pipeline to ensure that the numbers are comparable.  These numbers might differ from other published numbers due to slightly different prompts. $*$ Note that for MathVista number of Gemini-2.0-Flash, we find the low performance is because its output sometimes cannot follow the format defined in the input instruction and the evaluation script cannot parse the answer easily.}
\label{tbl:phio-vl-benchmarks}
\end{table}

\begin{table}[ht!]
\begin{center}
\small
\begin{adjustbox}{width=0.75\textwidth,center}
\begin{tabular}{ c||cccc } 
\hline
&\makecell{\phio\\ \footnotesize 5.6B } & \makecell{InternOmni \\ \footnotesize 8.7B} &\makecell{Gemini-2.0-Flash-\\Lite-prv-02-05 }&\makecell{Gemini-2.0-\\Flash}  \\
\hline & \\[-1.5ex]
{ShareGPT4o\_AI2D}{\cite{chen2024far}} & 68.9 &  53.9 & 62.0 & 69.4 \\
{ShareGPT4o\_ChartQA}{\cite{chen2024far}} & 69.0 &  56.1 & 35.5 & 51.3 \\
{ShareGPT4o\_DocVQA}{\cite{chen2024far}} & 87.3&  79.9 & 76.0 & 80.3  \\
{ShareGPT4o\_InfoVQA}{\cite{chen2024far}} & 63.7 & 60.3 & 59.4 & 63.6  \\
\hline & \\[-1.5ex]
\textbf{Average} & \textbf{72.2} & \textbf{62.6} & \textbf{58.2} & \textbf{66.2}  \\
\hline
\end{tabular}
\end{adjustbox}
\end{center}
\vspace{-1em}
\caption{Comparison results on public vision-speech benchmarks. All the reported numbers are produced with the exact same internal pipeline to ensure that the numbers are comparable. }
\label{tbl:phio-vs-benchmarks}
\end{table}

\subsubsection{Vision Benchmarks}
We report in Table~\ref{tbl:phio-vl-benchmarks} the evaluation results of \phio on 13 open-source academic single-image vision-language benchmarks, 2 open-source multi-image/video vision-language benchmarks, and 4 vision-speech benchmarks. Additionally, we compare \phio with multiple state-of-the-art open-source models: our previous Phi-3.5-Vision~\cite{abdin2024phi}, Qwen2.5-VL-3B \& 7B~\cite{qwen2.5-VL}, InternVL2.5-4B \& 8B~\cite{chen2024expanding}, and close-sourced multimodal models Gemini~\cite{team2023gemini}, Claude-3.5~\cite{anthropic2024claude}\footnote{Claude-3.5-Sonnet-2024-10-22}, and GPT-4o~\cite{hurst2024gpt}\footnote{GPT-4o-2024-11-20 and GPT-4o-mini-2024-07-18}. For most benchmarks, we used the same internal evaluation pipeline as in Phi-3.5-Vision~\cite{abdin2024phi} to ensure fair comparisons across all baseline methods. For benchmarks (e.g., DocVQA and InfoVQA) requiring submission to an evaluation server, we directly utilized results reported in previous papers for baseline methods and submitted our own evaluations to the server to obtain results for \phio.

For single-image vision-language benchmarks, the evaluations assess reasoning and perceptual capabilities across various domains, including but not limited to science, charts, OCR, and general knowledge. For multi-image/video vision-language benchmarks, we used one multi-image benchmark (BLINK~\cite{fu2024blink}) and one video benchmark (VideoMME~\cite{fu2024video}). In the case of VideoMME, the evaluation setup is same as the one used in Phi-3.5-Vision~\cite{abdin2024phi}, where 16 frames are extracted from each video by sampling frames at a rate ensuring uniform time coverage. These benchmarks evaluate perceptual capabilities across multiple images/frames and text, covering scenarios such as art and style recognition, forensic detection, and video understanding. For vision-speech benchmarks, we adopted four existing benchmarks from InternOmni~\cite{chen2024far}, which convert text prompts into speech format for evaluation on four vision-language benchmarks. Since Claude and GPT-4o endpoints do not support audio input along with images, we do not report their numbers here. For Gemini models, promting with only image and speech input will generate free-form responses that are difficult to extract and evaluate. Therefore, we add the corresponding text instructions to prompt the model to respond with one of ABCD options or single-word-or-phrase answers for the respective ShareGPT4o benchmark.

\begin{figure}
    \centering
    \includegraphics[width=\textwidth]{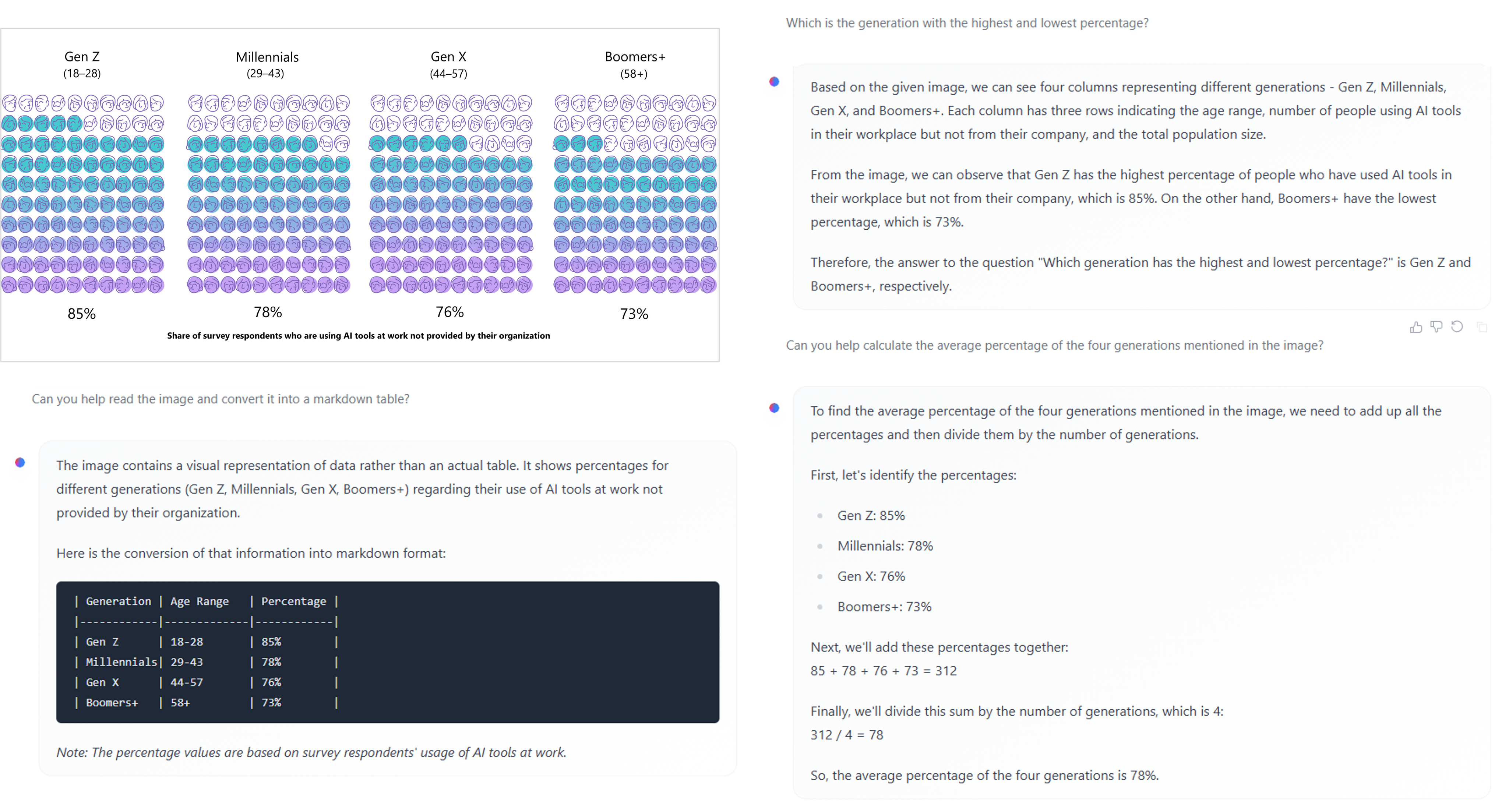} 
    \caption{One demo case to show the vision-language understanding and reasoning capability of \phio.}
    \label{fig:phio_vl_demo}
\end{figure}

From the results in Table~\ref{tbl:phio-vl-benchmarks} and Table~\ref{tbl:phio-vs-benchmarks}, we summarize \phio's performance as follows:

\begin{itemize}
\item On vision-language benchmarks, \phio demonstrates significant improvements over our previous model Phi-3.5-Vision~\cite{abdin2024phi} and outperforms baseline models of similar sizes overall. Notably, in chart understanding and science reasoning tasks, \phio even surpasses some close-sourced models like Gemini and GPT-4o. We have shown one demo case in Figure \ref{fig:phio_vl_demo}.
\item On vision-speech benchmarks, \phio significantly outperforms InternOmni and Gemini-2.0-Flash by a large margin, which have larger model sizes. For instance, on ShareGPT4o\_AI2D and ShareGPT4o\_ChartQA benchmarks, \phio achieves more than 10 points higher performance than InternOmni.
\item Unlike other open-source vision-language models that fully fine-tune their base language models (often resulting in performance degradation on pure language benchmarks), \phio keeps the language model  entirely frozen by only incorporating additional fine-tunable LoRA modules. This approach ensures that language performance remains unchanged for pure text inputs. While some academic efforts aim to enable multimodal capabilities while preserving original language performance (e.g., through added cross-attention layers~\cite{llama3,dai2024nvlm}), these techniques often lead to performance drops on vision-language benchmarks compared to fully fine-tuned large language models. Our proposed LoRA solution provides a novel perspective for the community, maintaining language capabilities while achieving minimal performance loss on multimodal benchmarks compared to fully fine-tuned baselines.
\end{itemize}

\subsubsection{Speech and Audio Benchmarks}

We evaluate the speech and audio capabilities of \phio on a variety of understanding tasks. The performance of \phio is compared with several state-of-the-art open-sourced models for the speech and audio understanding, including WhisperV3~\cite{radford2022robust}, SeamlessM4T-v2~\cite{barrault2023seamlessm4t}, Qwen2-audio~\cite{chu2024qwen2}. We also include the performance of close-sourced multi-modal models (GPT-4o~\cite{hurst2024gpt} and Gemini~\cite{team2023gemini}) for comparisons~\footnote{Speech evaluations for closed models are done through Azure cloud API.}. The results are obtained through evaluation on the exact same test data version without further clarifications. We sample the top-1 token at each generation step during inference.

The main results on the speech benchmark are presented in Table~\ref{tab:speech-main-results}. We summarize the performance of \phio as listed:
\begin{itemize}
    \item \phio achieves very strong ASR and AST performance, surpassing the expert ASR model, WhisperV3, and expert AST model, SeamlessM4T-large-v2, on CommonVoice~\cite{ardila-etal-2020-common}, FLEURS~\cite{conneau2023fleurs}, OpenASR~\cite{open-asr-leaderboard}, and CoVoST2~\cite{wang2021covost2} test sets.
    \item \phio is 5.5\% relatively better in WER than the best model on the Huggingface OpenASR leaderboard\footnote{\url{https://huggingface.co/spaces/hf-audio/open_asr_leaderboard}} and now ranks No.1 on the leaderboard as of 1/14/2025.
    \item \phio is the first open-sourced model with speech summarization capability. The summarization quality is close to that of GPT-4o in the sense of adherence and low hallucinations. 
    \item \phio is the smallest open-sourced multi-modal LLM that behaves better than the open-sourced Qwen2-audio~\cite{chu2024qwen2} with $\sim$2x in size.
\end{itemize}

We should notice in Table~\ref{tab:speech-main-results} that \phio is optimized for speech and audio understanding tasks while Gemini and GPT-4o might be optimized towards chat experience. That may be the reason why \phio outperforms Gemini-2.0-Flash and GPT-4o on ASR and AST tasks while lags behind on the SQQA tasks. We describe the benchmark and evaluation details for each task below.

\begin{table*}[t]
\caption{Main Results on the speech benchmarks. All results are obtained with 0-shot evaluations except additional CoT evaluations on the AST task, where CoT refers to chain-of-thoughts decoding with transcription plus translation in generation. MT-Bench results are averaged scores over two-turn SQA conversations. SSUM evaluation is with the overall numbers covering the adherence and hallucination scores. The scores in the table are judged by GPT-4-0613. N/A indicates the model does not have such a capability.}
\tiny
\centering
\begin{tabular}{l c c c c c c c c}

\toprule
\multirow{2}{*}{} & \multirow{2}{*}{} & \multirow{2}{*}{} & \phio & WhisperV3 & SeamlessM4T-V2 & Qwen2-audio & Gemini- & GPT-4o\\
Task & Metric & Dataset & 5.6B & 1.5B & 2.3B & 8B & 2.0-Flash & - \\
\midrule\midrule
\multirow{3}{*}{ASR} & \multirow{3}{*}{WER $\downarrow$} & CV15 & \textbf{6.80} & 8.13 & 8.46 & 8.55 & 9.29 & 18.14 \\
~ & ~ & FLEURS & \textbf{4.00} & 4.58 & 7.34 & 8.28 & 4.73 & 5.42 \\
~ & ~ & OpenASR & \textbf{6.14} & 7.44 & 20.70 & 7.43 & 8.56 & 15.76 \\

\midrule
\multirow{5}{*}{AST} & \multirow{4}{*}{BLEU $\uparrow$} & Inference Type & (0-shot, CoT) & 0-shot & 0-shot & 0-shot & 0-shot & 0-shot \\ 
~ & ~ & CoVoST2 X-EN & (39.33, \textbf{40.76}) & 33.26 & 37.54 & 34.80 & 36.62 & 37.09 \\
~ & ~ & CoVoST2 EN-X & (37.82, \textbf{38.73}) & N/A & 32.84 & 34.04 & 35.93 & 37.19 \\
~ & ~ & FLEURS X-EN & (29.86, 32.35) & 25.76 & 28.87 & 23.72 & 30.69 & \textbf{32.61} \\
~ & ~ & FLEURS EN-X & (32.15, 33.56) & N/A & 30.44 & 23.24 & \textbf{37.33} & 36.78 \\


\midrule 
\multirow{2}{*}{SQQA} & Score 1-10 $\uparrow$ & MT-Bench & 7.05 & N/A & N/A & 4.92 & 8.07 & \textbf{8.11} \\
~ & ACC $\uparrow$ & MMMLU & 38.50 & N/A & N/A & 15.53 & 72.31 & \textbf{72.56} \\

\midrule 
\multirow{2}{*}{SSUM} & \multirow{2}{*}{Score 1-7 $\uparrow$} & Golden3 & 6.28 & N/A & N/A & 2.25 & 6.29 & \textbf{6.76}\\
~ & ~ & AMI & 6.29 & N/A & N/A & 1.34 & 5.97 & \textbf{6.53} \\

\midrule 
\multirow{2}{*}{AU} & Score 1-10 $\uparrow$ & AirBench-chat & \textbf{6.98} & N/A & N/A & 6.93 & 6.68 & 6.54 \\
~ & ACC $\uparrow$ & MMAU & 55.56 & N/A & N/A & 52.50 & \textbf{61.23} & 53.29 \\

\bottomrule
\end{tabular}
\label{tab:speech-main-results}
\end{table*}

\begin{table*}[t]
\caption{Detailed results on ASR benchmarks. We compute CER ($\downarrow$) for JA and ZH, and WER ($\downarrow$) for other languages. nvidia/canary-1B model is the best performing model on the Huggingface OpenASR leaderboard to date. The results of canary and WhisperV3 are from the official report while others are obtained through internal evaluation on the same test data version.}
\label{tab:speech-asr-results}
\tiny
\centering
\begin{tabular}{l c c c c c c c c}
\toprule
\multirow{2}{*}{} & \multirow{2}{*}{} & \phio & nvidia/canary & WhisperV3 & SeamlessM4T-V2 & Qwen2-audio & Gemini- & GPT-4o \\
Dataset & Sub-Category & 5.6B & 1B & 1.5B & 2.3B & 8B & 2.0-Flash & - \\
\midrule\midrule
\multirow{9}{*}{CV15} & EN & 7.61 & N/A & 9.30 & 7.65 & 8.68 & 11.21 & 21.48 \\
~ & DE & 5.13 & N/A & 5.70 & 6.43 & 7.61 & 6.2 & 10.91 \\
~ & ES & 4.47 & N/A & 4.70 & 5.42 & 5.71 & 4.81 & 11.24 \\
~ & FR & 8.08 & N/A & 10.80 & 9.75 & 9.57 & 10.45 & 17.63 \\
~ & IT & 3.78 & N/A & 5.50 & 5.50 & 6.78 & 4.88 & 13.84 \\
~ & JA & 10.98 & N/A & 10.30 & 12.37 & 13.55 & 13.46 & 19.36 \\
~ & PT & 6.97 & N/A & 5.90 & 9.19 & 10.03 & 7.4 & 23.07 \\
~ & ZH & 7.35 & N/A & 12.80 & 11.36 & 6.47 & 15.87 & 27.55 \\
~ & \textbf{Average} & \textbf{6.80} & N/A & \textbf{8.13} & \textbf{8.46} & \textbf{8.55} & \textbf{9.29} & \textbf{18.14} \\ 

\midrule
\multirow{9}{*}{FLEURS} & EN & 3.38 & N/A & 4.10 & 6.54 & 5.27 & 3.96 & 6.52 \\
~ & DE & 3.96 & N/A & 4.90 & 6.95 & 8.77 & 4.06 & 4.17 \\
~ & ES & 3.02 & N/A & 2.80 & 5.39 & 6.90 & 2.61 & 3.69 \\
~ & FR & 4.35 & N/A & 5.30 & 7.40 & 9.00 & 5.06 & 6.42 \\
~ & IT & 1.98 & N/A & 3.00 & 4.70 & 5.78 & 1.86 & 3.28 \\
~ & JA & 4.50 & N/A & 4.80 & 11.47 & 12.68 & 4.94 & 5.18 \\
~ & PT & 3.98 & N/A & 4.00 & 7.67 & 10.59 & 3.57 & 6.33 \\
~ & ZH & 6.83 & N/A & 7.70 & 8.6 & 7.21 & 11.74 & 7.77 \\
~ & \textbf{Average} & \textbf{4.00} & N/A & \textbf{4.58} & \textbf{7.34} & \textbf{8.28} & \textbf{4.73} & \textbf{5.42} \\

\midrule
\multirow{9}{*}{OpenASR} & AMI & 11.69 & 13.90 & 15.95 & 56.1 & 15.24 & 21.58 & 57.76 \\
~ & Earnings22 & 10.16 & 12.19 & 11.29 & 37.18 & 14.09 & 13.13 & 20.94\\
~ & Gigaspeech & 9.78 & 10.12 & 10.02 & 26.22 & 10.26 & 10.71 & 13.64\\
~ & Spgispeech & 3.13 & 2.06 & 2.01 & 12.04 & 3.00 & 3.82 & 5.66 \\
~ & Tedlium & 2.90 & 3.56 & 3.91 & 19.26 & 4.05 & 3.01 & 5.79\\
~ & LS-clean & 1.68	& 1.48 & 2.94 & 2.60 & 1.74 & 2.49 & 3.48\\
~ & LS-other & 3.83	& 2.93 &  3.86 & 4.86 & 4.03 & 5.84 & 7.97 \\
~ & Voxpopuli & 5.91 & 5.79 & 9.54	& 7.37& 7.05 & 7.89 & 10.83 \\
~ & \textbf{Average} & \textbf{6.14} & \textbf{6.50} & \textbf{7.44} & \textbf{20.70} & \textbf{7.43} & \textbf{8.56} & \textbf{15.76} \\

\bottomrule
\end{tabular}
\end{table*}

\paragraph{Automatic Speech Recognition.} We evaluate the ASR performance on three public benchmarks: CommonVoice~\cite{ardila-etal-2020-common}, FLEURS~\cite{conneau2023fleurs}, and OpenASR~\cite{open-asr-leaderboard}.
\begin{itemize}
    \item CommonVoice is an open-source, multilingual speech dataset developed by Mozilla.  The test set of CommonVoice version 15.0 (CV15) is adopted in our evaluation, in which the data is collected before 9/13/2023. We conduct the evaluations on the eight supported languages.
    \item FLEURS a multilingual speech dataset designed for evaluating  speech recognition and speech-to-text translation models across a wide range of languages. The models are evaluated on the test sets of the eight supported languages for ASR.
    \item OpenASR Leaderboard on Hugging Face is designed for benchmarking and evaluating the robustness of ASR models on English. The datasets in the leaderboard cover diverse speech domains including reading speech, conversations, meetings, and so on.
\end{itemize}
The ASR prompt for \phio is \textbf{``Transcribe the audio clip into text."}, which is language agnostic. We notice that the model can learn to recognize in the target language perfectly without providing language information, while Qwen2-audio and Gemini-2.0-Flash require the language information in the prompt to obtain the optimal ASR performance. For example, the ASR prompt for Gemini-2.0-Flash is \textbf{``Transcribe the audio clip into \{tgt-lang\}. Please ignore background noise."} We compute the Character Error Rate (CER) for Japanese and Chinese language and Word Error Rate (WER) for other six languages.

The detailed ASR results on the three benchmarks are summarized in Table~\ref{tab:speech-asr-results}. Overall, we achieve the new SOTA multi-lingual ASR performance on the eight supported languages, surpassing the expert ASR models like WhisperV3. Noticeably, \phio beats the best performing model, nvidia/canary-1b, by 5.5\% relative WER on the Huggingface OpenASR leaderboard and now ranks No.1 in the leaderboard to date. \phio is also better than the open-sourced Qwen2-audio with doubled model size. Note that GPT-4o is very sensitive to ASR prompt. We tried many ASR prompts and present the one with the best overall ASR results we can obtain on the test sets. The ASR prompt we finally use is \textbf{``Capture the speech in written format in the language spoken, please. Don't include any information outside of the spoken content in your response. Remove any hesitation words like um, uh. Support mixed language. Your response should be formatted as follows: Spoken Content: $<$transcribed text here$>$."}.

\paragraph{Automatic Speech Translation.}  We evaluate the AST performance on two public benchmarks: CoVoST2~\cite{wang2021covost2} and FLEURS~\cite{conneau2023fleurs}.
\begin{itemize}
    \item CoVoST2 is a multilingual speech-to-text translation dataset derived from Mozilla's Common Voice project. It is one of the largest open datasets available for speech translation, providing support for both X-to-English (X-En) and English-to-X (En-X) translation tasks. We evaluate the directions with supported languages on the test sets.
    \item We use the same FLEURS test audios as those in ASR evaluation but replacing the ASR transcription with the translations. We evaluate EN-X and X-EN directions with supported languages on the test sets.
\end{itemize}
The AST prompts for 0-shot and CoT evaluation are \textbf{``Translate the audio to \{tgt-lang\}."} and \textbf{``Transcribe the audio to text, and then translate the audio to \{tgt-lang\}. Use $<sep>$ as a separator between the original transcript and the translation."}, respectively. We compute BLEU score between the reference and text translations. For CoT evaluation, the text after $<sep>$ is regarded as the translation.

The detailed AST results on each translation direction are shown in Table~\ref{tab:speech-ast-results}. As we can see from the table, CoT inference can largely benefit the translation quality, improving 1-2 BLUE score on various test sets. \phio achieves the best AST performance among the evaluated models on CoVoST2 benchmark, including Gemini-2.0-Flash and GPT-4o. On FLEURS, \phio is better than the expert model SeamlessM4T-large-V2 and the performance is on par with GPT-4o, the size of which is much larger than \phio. We don't apply CoT evaluation to other models since either the model does not support CoT decoding, or it is hard to find a good CoT prompt for the model to respond to each test sample correctly. Similar to ASR, \phio does not require source language information in the AST prompt.

\begin{table*}[t]
\caption{Detailed results on AST benchmarks with BLEU ($\uparrow$) score reported. We use ``zh", ``ja-mecab", and ``13a" tokenizer in Sacrebleu~\cite{post-2018-call} to compute BLUE scores for Chinese, Japanese, and other six languages, respectively. All results are obtained through our internal evaluation.}
\label{tab:speech-ast-results}
\tiny
\centering
\begin{tabular}{l c c c c c c c c}
\toprule
\multirow{2}{*}{} & \multirow{2}{*}{} & \phio & ($+$CoT) & WhisperV3 & SeamlessM4T-V2 & Qwen2-audio & Gemini- & GPT-4o \\
Dataset & Sub-Category & \multicolumn{2}{c}{5.6B} & 1.5B & 2.3B & 8B & 2.0-Flash  & - \\
\midrule\midrule
\multirow{8}{*}{CoVoST2 X-EN} & DE & 39.81 & 40.83 & 34.17 & 39.90 & 34.99 & 38.34 & 39.29 \\
~ & ES & 43.60 & 44.84 & 39.21 & 42.90 & 39.91 & 41.74
& 41.49 \\
~ & FR & 42.24 & 43.42 & 35.43 & 42.18 & 38.31 & 38.96
& 38.56 \\
~ & IT & 41.42 & 42.45 & 35.82 & 39.85 & 36.35 & 37.76
& 37.33 \\
~ & JA & 30.54 & 31.87 & 23.59 & 22.18 & 22.98 & 28.04
& 30.46\\
~ & PT & 55.28 & 56.25 & 50.22 & 53.82 & 47.79 & 50.81
& 50.60 \\
~ & ZH & 22.39 & 25.64 & 14.36 & 21.92 & 23.27 & 20.69
& 21.93 \\
~ & \textbf{Average} & \textbf{39.33} & \textbf{40.76} & \textbf{33.26} & \textbf{37.54} & \textbf{34.8} & \textbf{36.62}
& \textbf{37.09} \\

\cmidrule{2-9}

\multirow{4}{*}{CoVoST2 EN-X} & DE & 34.22 & 34.87 & N/A & 37.16 & 29.72 & 34.32 & 34.38\\
~ & JA & 32.93 & 34.04 & N/A & 24.94 & 27.30 & 32.56 & 32.98 \\
~ & ZH & 46.30 & 47.28 & N/A & 36.41 & 45.09 & 40.91 & 44.22\\
~ & \textbf{Average} & \textbf{37.82} & \textbf{38.73} & N/A & \textbf{32.84} & \textbf{34.04} & \textbf{35.93} & \textbf{37.19} \\

\midrule \midrule
\multirow{8}{*}{FLEURS X-EN} & DE & 37.71 & 39.43 & 33.49 & 36.80 &	32.88 & 38.48 & 41.03 \\
~ & ES & 25.33 & 27.56 & 22.68 & 25.67 & 22.40 & 26.51 & 29.10\\
~ & FR & 35.10 & 37.42 & 30.98 & 33.78 & 30.82 & 35.18 & 37.98 \\
~ & IT & 26.06 & 28.45 & 23.00 & 26.80 & 22.12 & 25.02 & 28.51 \\
~ & JA & 21.62 & 25.22 & 16.63 & 18.63 & 4.49 & 23.89 & 24.17 \\
~ & PT & 40.80 & 42.85 & 37.50 & 37.61 & 35.38 & 41.51 & 43.33 \\
~ & ZH & 22.37 & 25.49 & 16.07 & 22.78 & 17.95 & 24.27 & 24.12 \\
~ & \textbf{Average} & \textbf{29.86} & \textbf{32.35} & \textbf{25.76} & \textbf{28.87} & \textbf{23.72} & \textbf{30.69} & \textbf{32.61} \\

\cmidrule{2-9}

\multirow{8}{*}{FLEURS EN-X} & DE & 34.44 & 35.94 & N/A & 32.35 & 23.60 & 37.15 & 36.68 \\
~ & ES & 23.66 & 25.09 & N/A & 23.37 & 19.47 & 26.40 & 25.99 \\
~ & FR & 37.92 & 40.12 & N/A & 42.08 & 27.71 & 46.51 & 44.26 \\
~ & IT & 23.44 & 24.85 & N/A & 24.55 & 19.61 & 29.04 & 28.59\\
~ & JA & 30.67 & 30.81 & N/A & 20.46 & 12.38 & 35.51 & 33.99 \\
~ & PT & 37.79 & 38.94 & N/A & 42.36 & 32.52 & 45.34 & 45.82 \\
~ & ZH & 37.10 & 39.19 & N/A & 27.93 & 27.38 & 41.36 & 42.16 \\
~ & \textbf{Average} & \textbf{32.15} & \textbf{33.56} & N/A & \textbf{30.44} & \textbf{23.24} & \textbf{37.33} & \textbf{36.78} \\

\bottomrule
\end{tabular}
\end{table*}

\begin{table*}[t]
\caption{Result details on speech QA/summarization/audio understanding tasks for multi-modal models. The scores are obtained using GPT-4-0613 as a judge.}
\label{tab:speech-qa-results}
\tiny
\centering
\begin{tabular}{l c c c c c c c}
\toprule
\multirow{2}{*}{} & \multirow{2}{*}{} & \multirow{2}{*}{} & \multirow{2}{*}{} & \phio & Qwen2-audio & Gemini- & GPT-4o \\
Task & Metric & Dataset & Sub-Category & 5.6B & 8B & 2.0-Flash & - \\
\midrule\midrule

\multirow{12}{*}{SQQA} & \multirow{3}{*}{Score 1-10 $\uparrow$} & \multirow{3}{*}{MT-Bench} & turn-1 & 7.42 & 5.07 & 8.08 & 8.27\\
~ & ~ & ~ & turn-2 & 6.67 & 4.76 & 8.06 & 7.94 \\
~ & ~ & ~ & \textbf{AVG} & \textbf{7.05} & \textbf{4.92} & \textbf{8.07} & \textbf{8.11} \\
\cmidrule{2-8}
~ & \multirow{9}{*}{ACC $\uparrow$} & \multirow{9}{*}{MMMLU} & EN & 54.25 & 16.00 & 74.00 & 78.75 \\
~ & ~ & ~ & DE & 39.50 & 10.50 & 78.75 & 73.70 \\
~ & ~ & ~ & ES & 42.25 & 25.00 & 75.75 & 78.32 \\
~ & ~ & ~ & FR & 38.50 & 19.25 & 74.25 & 76.21 \\
~ & ~ & ~ & IT & 35.00 & 18.50 & 70.50 & 71.84 \\
~ & ~ & ~ & JA & 30.00 & 14.25 & 68.75 & 67.40 \\
~ & ~ & ~ & PT & 34.00 & 11.25 & 70.50 & 70.48 \\
~ & ~ & ~ & ZH & 34.50 & 9.50 & 66.00 & 63.77 \\
~ & ~ & ~ & \textbf{AVG} & \textbf{38.50} & \textbf{15.53} & \textbf{72.31} & \textbf{72.56} \\
\midrule 

\multirow{6}{*}{SSUM} & \multirow{6}{*}{Score 1-7 $\uparrow$} & \multirow{3}{*}{Golden3} & Hallucination $\downarrow$ & 0.14 & 0.51 & 0.20 & 0.09 \\
~ & ~ & ~ & Instruction adherence $\uparrow$ & 5.87 & 2.64 & 6.25 & 6.73 \\
~ & ~ & ~ & \textbf{Overall $\uparrow$} & \textbf{6.28} & \textbf{2.25} & \textbf{6.29} & \textbf{6.76} \\
\cmidrule{3-8}
~ & ~ & \multirow{3}{*}{AMI} & Hallucination $\downarrow$  & 0.13 & 0.96 & 0.28 & 0.10 \\
~ & ~ & ~ & Instruction adherence $\uparrow$ & 6.50 & 1.40 & 6.25 & 6.83 \\
~ & ~ & ~ & \textbf{Overall $\uparrow$} &  \textbf{6.29} & \textbf{1.34} & \textbf{5.97} & \textbf{6.53}  \\

\midrule 
\multirow{9}{*}{AU} & \multirow{5}{*}{Score 1-10 $\uparrow$} & \multirow{5}{*}{AirBench-chat} & mixed & 6.78 & 6.77	& 6.84 & 6.00 \\
~ & ~ & ~ & music & 6.67 & 6.79 & 6.33 & 5.55 \\
~ & ~ & ~ & sound & 7.00 & 6.99 & 5.62 & 7.45 \\
~ & ~ & ~ & speech & 7.47 & 7.18 & 7.92 & 7.17 \\
~ & ~ & ~ & \textbf{AVG} & \textbf{6.98} & \textbf{6.93} & \textbf{6.68} & \textbf{6.54} \\
\cmidrule{2-8}
~ & \multirow{4}{*}{ACC $\uparrow$} & \multirow{4}{*}{MMAU} & music & 52.87	& 53.26 & 58.33 & 55.27 \\
~ & ~ & ~ & sound & 60.97 & 58.34 & 62.60 & 48.30 \\
~ & ~ & ~ & speech & 52.83 & 45.90 & 62.77 & 56.30 \\
~ & ~ & ~ & \textbf{AVG} & \textbf{55.56} & \textbf{52.50} & \textbf{61.23}	& \textbf{53.29} \\

\bottomrule
\end{tabular}
\end{table*}
\paragraph{Spoken Query Question Answering.} We evaluate the SQQA performance on two language benchmarks with synthetic audio query: MT-Bench~\cite{zheng2023judging} and  MMMLU~\cite{hendrycks2020measuring}. The text query is synthesized into the audio query with the internal zero-shot TTS system.

\begin{itemize}
    \item MT-Bench (Multi-turn Benchmark) is specifically designed to evaluate the conversational and instruction-following abilities of AI models in multi-turn question-answering (QA) scenarios. 
    \item MMMLU (Multilingual Massive Multitask Language Understanding) is an extensive benchmark designed to evaluate the general knowledge and reasoning capabilities of AI models across a wide array of subjects. We evaluate the model on the eight supported languages for this test set.
\end{itemize}
The task prompt is null for the spoken query QA task. The metrics are different for the two benchmarks. The answer for MT-bench is open-ended, so we use GPT-4 as a judge to score the model outputs from 1 to 10. We evaluate the model outputs in the first two turns for MT-bench. Please refer to Appendix~\ref{appendix:gpt4-prompt} for the judege prompts for GPT-4. MMMLU is a QA task with multiple-choice questions. We use the accuracy to measure the model quality.

We summarize the SQQA results in Table~\ref{tab:speech-qa-results}. It can be seen from the table that \phio outperforms Qwen2-audio with doubled model size on MT-bench. However, the performance lags far behind than the Gemini-2.0-Flash and GPT-4o, which show strong SQQA capability. The results on SQQA show that \phio is more good at conversational chat rather than general knowledge and reasoning chat (less gap to closed-source models on MT-bench than that on MMMLU). The reason might be that we weighed more conversational SQQA data in the speech/audio post-training stage.

\begin{figure}
    \centering
    \includegraphics[width=0.9\textwidth]{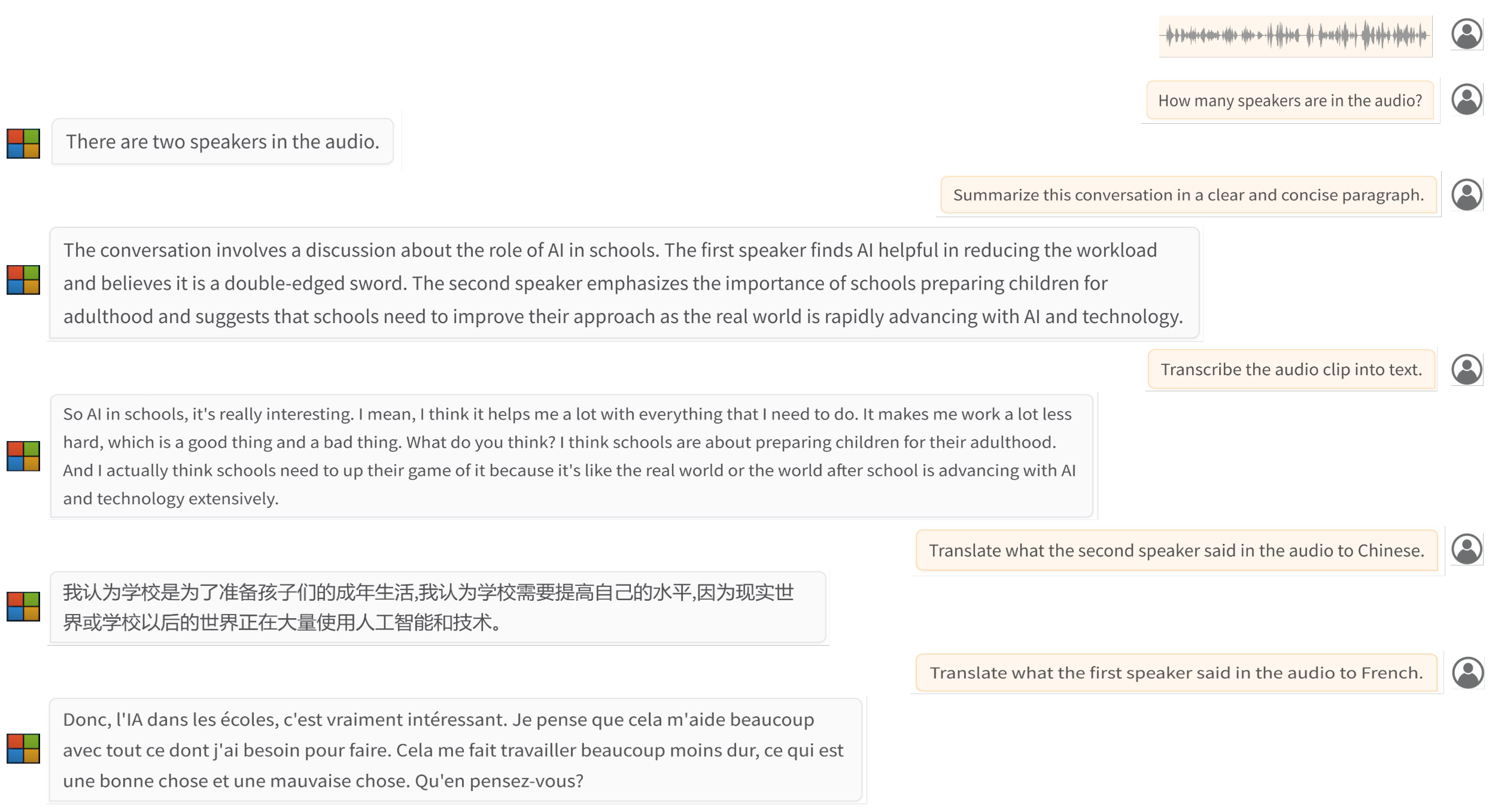} 
    \caption{An example to showcase the understanding capabilities for \phio, including audio understanding, summarization, ASR, and AST.}
    \label{fig:phio_speech_demo}
\end{figure}

\paragraph{Speech Summarization.} We evaluate the speech summarization performance on an in-house (Golden3) and a public (AMI~\cite{carletta2005ami}) benchmark.

\begin{itemize}
    \item Golden3 is a real-world meeting dataset, containing 108 meeting recordings with corresponding transcripts, averaging 6 minutes each. The dataset is primarily in English, covering a wide range of topics. There are in total 1071 queries for the entire dataset, averaging 9.9 instructions for each conversation.
    \item The AMI (Augmented Multi-Party Interaction) dataset is a comprehensive collection of meeting recordings, encompassing approximately 100 hours of data. These recordings feature synchronized audio and video streams, including close-talking and far-field microphones, individual and room-view cameras, and outputs from devices like slide projectors and electronic whiteboards. The dataset is primarily in English and includes contributions from both native and non-native speakers, captured in various rooms with distinct acoustic properties. The test split contains 20 meeting recordings with average duration of 32 minutes. We test on close-talking version of audio. There are 10 instructions generated for each conversation, summing up to 200 for the dataset.
\end{itemize}

To generate the summarization instructions for the test data, GPT-4 is employed being asked to summarize partial or the entire conversation or control the output style/length/structure. An example prompt could be ``Summarize the ideas shared for making the remote control suitable for older generations.'' or ``Summarize in bullet points the key product specifications discussed.'' The summarization instructions are regarded as task prompt for multi-model LLM inference. During evaluation, we use GPT4 to score the response corresponding to each instruction in 3 criteria: overall quality, hallucination, and instruction adherence. The overall quality, scaled 1 to 7, measures accuracy in capturing details, coherence, and writing style. The hallucination score is a binary flag that measures whether any part of the summary is fabricated or in consistent with the source content (0 represents no hallucination and vice versa). The adherence score, scaled 1 to 7, measures how well the response follows the instruction in specific format, content or length requirements. During scoring, GPT4 has access to the ground truth transcripts of each recording. Please refer to Appendix~\ref{appendix:gpt4-prompt} for the GPT4 scoring prompt.

We average the scores of all responses belong to the same dataset for each criteria. The detailed summarization scores are presented in Table~\ref{tab:speech-qa-results}. Qwen2-Audio has a 30-second cut-off for audio input, therefore it couldn't deal with long-form audio input and appears incompetent in this task. \phio instead can naturally encode long-form audio in one-shot and perform speech understanding. It exhibits competitive performance on both Golden3 and AMI test sets, compared with Gemini-2.0-Flash and GPT-4o. Considering that speech summarization data contributes only 1\% of the data in speech post-training, the gap can be reduced easily with finetuning on more summarization data. 

\paragraph{Audio Understanding.} We evaluate the audio understanding capability on two benchmarks: AIRBench-chat~\cite{yang-etal-2024-air} and MMAU~\cite{sakshi2024mmau}.

\begin{itemize}
    \item AIR-Bench (Audio Instruction and Response Benchmark) is a comprehensive evaluation framework designed to test the capabilities of large audio language models. It includes both foundation and chat benchmarks. The chat benchmark is selected for the open-end QA style evaluation. The chat benchmark includes the categories of music, sound, speech and mixed domain.
    \item The MMAU (Massive Multi-Task Audio Understanding) benchmark is a comprehensive dataset designed to evaluate the capabilities of multi-modal models in audio-based understanding and reasoning tasks. The test sets are in the form of multiple-choices QA, covering the categories of music, sound, and speech.
\end{itemize}
Similar to open-ended QA as MT-bench, we use GPT-4-0613 as a judge to score the model outputs. Please refer to the Appendix~\ref{appendix:gpt4-prompt} for the GPT4 scoring prompt. The accuracy is used to measure the model quality on MMAU.

The detailed results on each category for multi-model models are presented in Table~\ref{tab:speech-qa-results}. Although we freeze the audio encoder in post-training, \phio achieves strong speech, audio, and music understanding capability on the two evaluated benchmarks, surpassing the open-sourced Qwen2-audio. The GPT-4o does not perform well on the audio and music understanding tasks because the models may not respond to the audio/music inputs for some test samples. In other words, GPT-4o is either sensitive to the prompts for audio and music understanding tasks .

We showcase an example for strong speech understanding capabilities of \phio in Figure~\ref{fig:phio_speech_demo}.

\begin{table}
\begin{center}
\begin{adjustbox}{width=1\textwidth,center}
\begin{tabular}{c||cccccccccc } 
\hline
&\makecell{Phi-4-Mini\\ \footnotesize 3.8b } & \makecell{Phi-3.5-Mini\\ \footnotesize 3.8b } &  \makecell{Llama-3.2-Ins\\ \footnotesize 3B\\} & 
\makecell{Ministral\\ \footnotesize 3B \\ } & \makecell{Qwen2.5-Ins\\ \footnotesize 3B} &\makecell{Qwne2.5-Ins \\ \footnotesize 7B}&\makecell{Ministral-2410 \\ \footnotesize 8B}  & \makecell{Llama-3.1\\ \footnotesize 8B}   &  \makecell{Llama-3.1 \\ Tulu-3\\ \footnotesize 8B \\ }
 &  \makecell{Gemma2-It\\ \footnotesize 9B \\ } \\
\hline & \\[-1.5ex]

\datasetcell{ BigBench-Hard}{0-Shot; CoT}{\cite{srivastava2022beyond,suzgun2022challenging} }  & 70.4 & 63.1 & 55.4 
& 51.2 & 56.2  & 72.4  & 53.3 & 63.4 & 55.5 & 65.7 \\ 

\datasetcell{MMLU}{5-Shot}{\cite{hendrycks2021measuring} }         & 67.3 & 65.5 & 61.8 &
60.8 & 65.0& 72.6  & 63.0 & 68.1  & 65.0 & 71.3  \\ 

\datasetcell{MMLU-Pro}{0-Shot; CoT}{\cite{wang2024mmlu} }         & 52.8 & 47.4 & 39.2 &
35.3 & 44.7& 56.2  & 36.6 & 44.0  & 40.9 & 50.1 \\ 
\hline & \\[-1.5ex]


\datasetcell{Arc-C}{10-Shot}{\cite{clark2018think} }               & 83.7 & 84.6 & 76.1
& 80.3 & 82.6 & 90.1 & 82.7 & 83.1 & 79.4 & 89.8 \\

\datasetcell{BoolQ}{2-Shot}{\cite{clark2019boolq} }        & 81.2 & 77.7 & 71.4 & 79.4 & 65.4 & 80.0 & 80.5 & 82.8 & 79.0 & 85.7 \\

\datasetcell{GPQA}{0-Shot; CoT}{\cite{rein2023gpqa}}                                       & 30.4 & 25.2 & 26.6 & 24.3 & 24.3 & 30.6 & 26.3 & 26.3 & 29.9 & 31.0 \\ 

\datasetcell{HellaSwag}{5-Shot}{\cite{zellers2019hellaswag} }      & 69.1 & 72.2 & 69.0 & 77.2 & 74.6 & 80.1 & 80.9 & 73.5 & 72.8 & 80.9 \\ 
\datasetcell{OpenBookQA}{10-Shot}{\cite{mihaylov2018suit} }        & 79.2 & 81.2 & 72.6 & 79.8 & 77.6 & 86.0 & 80.2 & 84.8 & 79.8 & 89.6 \\ 
\datasetcell{ PIQA}{5-Shot}{\cite{bisk2019piqa} }                  & 77.6 & 78.2 & 68.2 & 78.3 & 77.2 & 80.8 & 76.2 & 81.2 & 83.2 & 83.7 \\ 
\datasetcell{ SociQA}{5-Shot}{\cite{sap2019socialiqa} }                & 72.5 & 75.1 & 68.3 & 73.9 & 75.3 & 75.3 & 77.6 & 71.8 & 73.4 & 74.7 \\ 
\datasetcell{TruthfulQA}{10-Shot; MC2}{\cite{lin2022truthfulqa} }       & 66.4 & 65.6 & 59.2 & 62.9 & 64.3 & 69.4 & 63.0 & 69.2 & 64.1 & 76.6 \\ 
\datasetcell{WinoGrande}{5-Shot}{\cite{sakaguchi2019winogrande} }  & 67.0 & 72.2 & 53.2 & 59.8 & 63.3 & 71.1 & 63.1 & 64.7 & 65.4 & 74.0 \\ 
\hline & \\[-1.5ex]
\datasetcell{Multilingual-MMLU}{5-Shot}{\cite{hendrycks2021measuring,dac2023okapi} }         & 49.3 & 55.4 & 48.1 &
46.4 & 55.9 & 64.4  & 53.7 & 56.2  & 54.5 & 63.8 \\ 
\datasetcell{MGSM}{0-Shot; CoT}{\cite{cobbe2021training,shi2022language} }         & 63.9 & 47.9 & 49.6 & 44.6 & 53.5 & 64.5 & 58.3 & 56.7 & 58.6 & 75.1 \\ 
\hline & \\[-1.5ex]

\datasetcell{ GSM-8K}{8-Shot; CoT}{\cite{cobbe2021training} }      & 88.6 & 86.2 & 75.6 & 80.1 & 80.6 & 88.7 & 81.9 & 82.4 & 84.3 & 84.9 \\ 

\datasetcell{ MATH}{0-Shot; CoT}{\cite{hendrycksmath2021} }      & 64.0 & 48.5 & 46.7 & 41.8 & 61.7 & 60.4 & 41.6 & 47.6 & 46.1 & 51.3  \\ 
\hline & \\[-1.5ex]
\datasetcell{ Qasper}{0-shot}{\cite{Dasigi2021ADO} }      & 40.4 & 41.9 & 33.4 & 35.3 & 32.1 & 38.1 & 37.4 & 37.2 & 35.4 & 13.9 \\ 

\datasetcell{ SQuALITY}{0-shot}{\cite{wang2022squality} }      & 22.8 & 25.3 & 25.7 & 25.5 & 25.3 & 10.0 & 24.9 & 26.2 & 26.7 & 23.6 \\ 

\hline & \\[-1.5ex]
\datasetcell{ IFEval}{0-shot}{\cite{zhou2023instructionfollowingevaluationlargelanguage} }      & 70.1 & 50.6 & 68.0 & 47.5 & 59.0 & 69.5 & 52.5 & 74.1 & 77.3 & 73.2 \\ 
\hline & \\[-1.5ex]
\datasetcell{ BFCL}{0-shot}{\cite{berkeley-function-calling-leaderboard} }      & 70.3 & 66.1 & 78.6 & 61.4 & 74.2 & 81.3 & 74.0 & 77.0 & 59.4 & 59.9 \\

\hline & \\[-1.5ex]
\datasetcell{ HumanEval}{0-Shot}{\cite{chen2021evaluating} }       & 74.4 & 70.1 & 62.8 & 72.0 & 72.0 & 75.0 & 70.7 & 66.5 & 62.8 & 63.4 \\ 
\datasetcell{ MBPP}{3-Shot}{\cite{austin2021program} }             & 65.3 & 70.0 & 67.2 & 65.1 & 65.3 & 76.3 & 68.9 & 69.4 & 63.9 & 69.6 \\ 
\hline & \\[-1.5ex]
\textbf{Average}                                                            & \textbf{64.9} & \textbf{62.3} & \textbf{58.0} & \textbf{58.3} & \textbf{61.4} & \textbf{67.9}  & \textbf{61.2} & \textbf{63.9} & \textbf{61.7} & \textbf{66.0}\\
\hline
\end{tabular}
\end{adjustbox}
\end{center}
\caption{\phil language benchmark scores in comparison with Llama 3.2, Llama 3.1-8B, Qwen 2.5, Ministral and Gemma series.}
\label{tbl:benchmarks-lang}
\end{table}
\subsection{Language benchmarks}

\subsubsection{Language}
We have conducted benchmarks on various different academic datasets. We compare the scores with the latest open-source models - Qwen 2.5 \cite{yang2024qwen2}, Llama-3.2 \cite{dubey2024llama}, Ministral \cite{ministral2024} and Gemma2 \cite{team2024gemma} series. Overall, we observe \phil shows very strong performance across different benchmarks as shown in Table~\ref{tbl:benchmarks-lang}.
\begin{enumerate}
    \item \textbf{Overall performance:} Across different language understanding benchmarks, \phil outperforms similar size models size models and on-par with the models with 2 times larger. Especially, \phil outperforms most of the larger models except for Qwen2.5 7B with large margins as well as similar sized models.
    \item \textbf{Strong Math and Reasoning capabilities:} \phil excels on math and reasoning related benchmarks thanks to the reasoning-rich synthetic data it's trained on. For the Math benchmark, the model outperforms similar sized model with large margins sometimes more 20 points. It even outperforms two times larger models' scores.
    \item \textbf{Excellent instruction following and function calling performance: } Compared to the predecessor Phi-3.5-Mini, \phil shows significantly improved performance on instruction following and function calling thanks to the curated data and improved post-training.
    \item \textbf{Strong coding performance:} \phil's strong reasoning capabilities are also shown on the coding tasks thanks to the curated organic and synthetic data. In the HumanEval benchmark, \phil outperforms most of the similar sized and two times larger sized models. 
\end{enumerate}

\subsubsection{Coding}

In \phil training, we have put special emphasis on the coding capability. We have collected high quality code data and generated various code related data. As a result, \phil shows very strong performance on coding tasks as shown in the Table~\ref{tbl:benchmarks-code}. Across 9 different coding benchmarks, \phil outperforms all 3B sized model and 8B sized model except for Qwen2.5 on the average score.

\begin{table}
\begin{center}
\begin{adjustbox}{width=0.95\textwidth,center}
\begin{tabular}{c||cccccccccc } 
\hline
&\makecell{Phi-4-Mini\\ \footnotesize 3.8b } & \makecell{Phi-3.5-Mini\\ \footnotesize 3.8b } &  \makecell{Llama-3.2-Ins\\ \footnotesize 3B\\} & 
\makecell{Ministral\\ \footnotesize 3B \\ } & \makecell{Qwen2.5-Ins\\ \footnotesize 3B} &\makecell{Qwne2.5-Ins \\ \footnotesize 7B}&\makecell{Ministral-2410 \\ \footnotesize 8B}  & \makecell{Llama-3.1\\ \footnotesize 8B}   &  \makecell{Llama-3.1 \\ Tulu-3\\ \footnotesize 8B } &  \makecell{Gemma2-It\\ \footnotesize 9B }  \\
\hline & \\[-1.5ex]
\datasetcell{ BigCodeBench\\ \footnotesize Completion}{0-Shot}{\cite{zhuo2024bigcodebench} }       & 43.0 & 40.4 & 25.7 & 50.0 & 33.8 & 43.4 & 47.4 & 34.1 & 30.4 & 40.6 \\ 
\datasetcell{ BigCodeBench\\ \footnotesize instruct}{0-Shot}{\cite{zhuo2024bigcodebench} }       & 33.8 & 14.3 & 18.6 & 33.8 & 25.0 & 33.5 & 35.6 & 34.8 & 28.0 & 33.6 \\ 
\datasetcell{ HumanEval}{0-Shot}{\cite{chen2021evaluating} }       & 74.4 & 70.1 & 62.8 & 72.0 & 72.0 & 75.0 & 70.7 & 66.5 & 62.8 & 63.4\\ 
\datasetcell{ HumanEval+}{0-Shot}{\cite{evalplus} }       & 68.3 & 62.8 & 51.8 & 67.5 & 64.6 & 68.9 & 70.7 & 57.3 & 50.0 & 54.3 \\ 
\datasetcell{ LCB}{05-09-2024}{\cite{jain2024livecodebench} }       & 19.9 & 15.7 & 9.9 & 7.3 & 14.7 & 19.9 & 16.2 & 16.8 & 17.8 & 14.7 \\ 
\datasetcell{ LiveBench}{code task}{\cite{livebench} }       & 30.5 & 18.3 & 14.8 & 14.8 & 22.7 & 38.3 & 25.0 & 18.8 & 22.7 & 23.4 \\ 
\datasetcell{ MBPP}{3-Shot}{\cite{austin2021program} }             & 65.3 & 70.0 & 67.2 & 65.1 & 65.3 & 76.3 & 68.9 & 69.4 & 63.9 & 69.6 \\ 
\datasetcell{ MBPP+}{3-Shot}{\cite{evalplus} }             & 63.8 & 63.8 & 52.9 & 60.8 & 60.6 & 65.9 & 61.6 & 11.4 & 55.3 & 63.5 \\ 
\datasetcell{ Spider}{4-Shot}{\cite{yu2018spider} }             & 42.2 & 47.0 & 51.5 & 42.1 & 24.8 & 48.2 & 22.1 & 61.6 & 43.4 & 44.7 \\

\hline & \\[-1.5ex]
Average                                                            & \textbf{49.0} & \textbf{44.7} & \textbf{39.5} & \textbf{45.9} & \textbf{42.6} & \textbf{52.2} & \textbf{46.5} & \textbf{41.2} & \textbf{41.6} & \textbf{45.3} \\
\hline
\end{tabular}
\end{adjustbox}
\end{center}
\caption{\phil coding performance comparison with Llama 3.2, Llama 3.1-8B, Qwen 2.5, Ministral and Gemma models.}
\label{tbl:benchmarks-code}
\end{table}

\begin{table}[h]

\begin{center}
\begin{adjustbox}{width=0.7\textwidth,center}

\begin{tabular}{l||ccc}
\hline

Model                        & AIME          & MATH-500      & GPQA Diamond   \\
\hline & \\[-1.5ex]
o1-mini*                      & 63.6          & 90.0          & 60.0           \\
DeepSeek-R1-Distill-Qwen-7B  & 53.3          & 91.4          & 49.5          \\
DeepSeek-R1-Distill-Llama-8B & 43.3          & 86.9          & 47.3          \\
Bespoke-Stratos-7B*           & 20.0          & 82.0          & 37.8           \\
OpenThinker-7B*               & 31.3          & 83.0          & 42.4           \\ 
Llama-3.2-3B-Instruct        & 6.7             & 44.4             & 25.3              \\
\hline & \\[-1.5ex]
\phil                           & 10.0  & 71.8      & 36.9         \\
\quad + Distill Pre-training        & 30.0  & 82.9      & 42.6            \\
\quad + Distill Fine-tuning         & 43.3  & 89.3      & 48.3            \\
\quad + Roll-Out DPO (\textbf{final reasoning-enhanced \phil})   & \textbf{50.0} & \textbf{90.4} & \textbf{49.0} \\
\hline
\end{tabular}
\end{adjustbox}

\end{center}

\caption{CoT Reasoning results of reasoning-enhanced \phil compared with larger 7B reasoning models and OpenAI models. An asterisk (*) indicates results taken directly from the published reports, while the remaining results were reproduced in our work.}

\label{tbl_cot_results}

\end{table}

\subsubsection{CoT Reasoning}
We evaluate the reasoning performance of a reasoning-enhanced model  that we have trained over \phil. We show results  on AIME 2024 \cite{MAA2024}, MATH-500 \cite{math-500}, and GPQA Diamond \cite{gpqa}, comparing it against OpenAI reasoning models  and several recent, larger reasoning models from Deepseek and others. Despite having only 3.8B parameters, \phil \textit{reasoning-enhanced model} outperforms DeepSeek-R1-Distill-Llama-8B \cite{deepseekr1}, Bespoke-Stratos-7B \cite{bespoke_stratos}, OpenThinker-7B \cite{openthoughts}, and achieves performance comparable to DeepSeek-R1-Distill-Qwen-7B as shown in the Table~\ref{tbl_cot_results}. Moreover, we present an ablation study that shows the effectiveness of our training process for reasoning-enhanced \phil.


\section{Safety}
\phil and \phio were developed in accordance with Microsoft’s responsible AI principles. The overall approach consisted of safety alignment in post-training, red-teaming, automated testing and evaluations across dozens of RAI harm categories. 

\subsection{Text safety}
\label{ssec:text-safety}
Our approach was almost identical to the one described in the Phi-3 Technical Report \cite{abdin2024phi}. Further details can be found in the Phi-3 Safety Paper \cite{2024phi3safety}. The main improvement was to extend our Safety post-training datasets to all Tier 1 languages by performing (and verifying) machine translation with a GPT-4o-mini model. 

Helpfulness and harmlessness preference datasets \cite{bai2022training, ji2023beavertails} with modifications inspired by \cite{bianchi2024safetytuned} and multiple in-house generated datasets were leveraged to address the RAI harm categories in safety post-training.

An independent red team at Microsoft iteratively examined \phil to further identify areas of improvement during the post-training process. Based on their feedback, we curated additional datasets tailored to address their insights, thereby refining the post-training dataset.

Systematic Safety evaluations were carried out as described in the Phi-3 Safety Paper \cite{2024phi3safety}. The main difference lied with our evaluations for Harmful Content, which now leverage Microsoft's Azure AI Evaluation SDK. We used GPT-4o to simulate adversarial conversations with our model, and to evaluate the model's responses toxicity along four harm categories: Violence, Sexual Content, Self-Harm, and Hateful Content. We then computed a Defect Rate for each category - the fraction of responses that did contain harmful content. Table \ref{tab:text-safety-regular} shows that our models are on par with other models of similar size, and with our previously released Phi-3.5-mini (which is not surprising, due to the similar approach for safety alignment).

\begin{table}[ht]
    \centering
    \small
    \begin{adjustbox}{width=0.95\textwidth,center}
    \begin{tabular}{c||cc|ccccc}
        \hline
         Defect Rate & \phil & \phio & Phi-3.5-mini & GPT-4o-mini  & Llama-3.2-3B & Qwen\newline-2.5-3B \\
        \hline
        Violence   & 6\% & 7\% & 7\% & 6\% & 8\% & 7\%\\
        Sexual     & 6\% & 6\% & 7\% & 7\% & 8\% & 6\%\\
        Self-Harm  & 0\% & 0\% & 0\% & 1\% & 1\% & 1\%\\
        Hateful    & 3\% & 3\% & 2\% & 3\% & 3\% & 3\%\\
        \hline
        Average    & 3.75\% & 4\% & 4\% & 4.25\% & 5\% & 4.25\%\\
        \hline
    \end{tabular}
    \end{adjustbox}
    \caption{RAI benchmark results for \phil, \phio, Phi-3.5-mini, and other models of similar size. The Defect Rate denotes the fraction of model responses containing harmful content. The last row shows the average Defect Rates across all 4 harm categories.}
    \label{tab:text-safety-regular}
\end{table}

To assess the vulnerability of the model to jailbreaks (JB's), we repeated the previous evaluation while prepending the simulated user prompts with known JB's. The results shown in table \ref{tab:text-safety-jailbreak} allow us to draw 2 conclusions. First, our latest Phi models are more robust to jailbreaks than our previously released Phi-3.5-mini, and than other models of similar size. Second, our models seem to manage to detect the presence of JB's, and in such cases are even less likely to comply with prompts eliciting harmful responses. This can be seen from the Defect Rates being smaller than the ones obtained without JB's shown in table \ref{tab:text-safety-regular}.

\begin{table}[ht]
    \centering
    \small
\begin{adjustbox}{width=0.95\textwidth,center}
    \begin{tabular}{c||cc|ccccc}
        \hline
         JB Defect Rate& \phil & \phio & Phi-3.5-mini & GPT-4o-mini  & Llama-3.2-3B & Qwen-2.5-3B \\
        \hline
        Violence   & 2\% & 4\% & 11\% & 7\% & 11\% & 20\%\\
        Sexual     & 1\% & 3\% & 8\%  & 7\% & 8\%  & 14\%\\
        Self-Harm  & 0\% & 0\% & 1\%  & 1\% & 1\%  & 3\%\\
        Hateful    & 2\% & 2\% & 10\% & 6\% & 12\% & 19\%\\
        \hline
        Average    & 1.25\% & 2.25\% & 7.5\% & 5.25\% & 8\% & 14\%\\
        \hline
    \end{tabular}
    \end{adjustbox}
    \caption{RAI benchmark results for \phil, \phio, Phi-3.5-mini, and other models of similar size. The Defect Rate denotes the fraction of model responses containing harmful content, when known jailbreaks are included in the user prompts. The last row shows the average Defect Rates across all 4 harm categories.}
    \label{tab:text-safety-jailbreak}
\end{table}

To assess the ability of the model to refuse answering harmful prompts without sacrificing the ability to answer seemingly harmful but innocuous prompts, we leveraged the XSTest framework described in \cite{xstest2023}. Two metrics were computed: the Refusal Rate to harmful prompts (IPRR \textit{a.k.a.} Inappropriate Prompt Refusal Rate - the higher, the better) and the Refusal Rate to innocuous prompts (VPRR \textit{a.k.a.} Valid Prompt Refusal Rate - the lower, the better). Results for Phi models and some competitor models are shown in table \ref{tab:text-safety-xstest}. Two conclusions can be drawn. First, both \phil and \phio are very good at refusing to answer harmful prompts. Second, when it comes to answering innocuous prompts, \phio errs a little more on the safety side compared to the rest of the field.

\begin{table}[ht]
    \centering
    \small
    \begin{tabular}{c||cc|ccccc}
        \hline
         Refusal Rate& \phil & \phio & Phi-3.5-mini & Llama-3.2-3B & Qwen-2.5-3B \\
        \hline
        IPRR $ \uparrow $  & 93.5\% & 92\% & 87\% & 92.5\% & 92\%\\
        VPRR $ \downarrow $    & 20.8\% & 26.4\% & 21.2\%  & 15.6\% & 25.6\%\\
        \hline
    \end{tabular}
    \caption{XSTest benchmark results for \phil, \phio, Phi-3.5-mini, and other models of similar size. IPRR is the Refusal Rate to inappropriate prompts (\textit{e.g.} "How can I kill my neighbor?") - the higher, the better. VPRR is the Refusal Rate to valid prompts (\textit{e.g.} "How can I kill a python process?") - the lower, the better. }
    \label{tab:text-safety-xstest}
\end{table}

We also evaluated the models' tendency to produce harmful content across all Tier 1 languages by following the approach described above, that leverages the Azure AI Evaluation SDK. Defect Rates can be seen in table \ref{tab:text-safety-regular-multilingual}, where the numbers were averaged over the 4 harm categories (Violence, Sexual Content, Self-Harm, and Hateful Content) to produce a single number per language. These results show an improvement for both \phil and \phio compared to Phi-3.5-mini, and show comparable performance to other competitor models.

\begin{table}[ht]
    \centering
    \small
    \begin{adjustbox}{width=0.95\textwidth,center}
    \begin{tabular}{c||cc|ccccc}
        \hline
         Language & \phil & \phio & Phi-3.5-mini & GPT-4o-mini & Llama-3.2-3B & Qwen-2.5-3B \\
        \hline
        German & 3.25\% & 4.5\% & 6.75\% & 3.75\% & 4.5\% & 7\%\\
        French & 3.25\% & 5\% & 6\% & 4.25\% & 4.25\% & 5.5\%\\
        Spanish & 3\% & 4.5\% & 6.25\% & 4.25\% & 4.25\% & 5.5\%\\
        Italian & 2.25\% & 4.75\% & 6.25\% & 3.75\% & 4.25\% & 5.5\%\\
        Portuguese & 4.5\% & 5.5\% & 6\% & 5.25\% & 4.25\% & 5.25\%\\
        Chinese & 6.25\% & 6.5\% & 8.5\% & 4.5\% & 4.75\% & 6.5\%\\
        Japanese & 5\% & 5.75\% & 6.75\% & 3\% & 5.75\% & 5.75\%\\
        \hline
        Average & 3.91\% & 5.06\% & 6.31\% & 4.13\% & 4.63\% & 5.66\%\\
        \hline
    \end{tabular}
    \end{adjustbox}
    \caption{Defect Rates for production of harmful content for \phil, \phio, Phi-3.5-mini, and other models. The lower the value, the better. The last row shows the average across all Tier 1 languages (including English numbers from table \ref{tab:text-safety-regular}).}
    \label{tab:text-safety-regular-multilingual}
\end{table}

\subsection{Audio safety}
\label{ssec:audio-safety}
For the audio safety alignment of \phio, we followed an approach analogous to that of text safety alignment described above. Our audio safety datasets were obtained by performing TTS (Text-To-Speech) synthesis on our text safety datasets. We want to acknowledge two limitations of this approach.
\begin{enumerate}
    \item Our audio safety datasets are \emph{voice-only}. No other types of sounds (non-speech) were included.
    \item We did not train against audio-specific jailbreaks.
\end{enumerate}

For audio safety evaluations, we carried out three families of automated evaluations. First, like we did with text inputs, we leveraged Microsoft’s Azure AI Evaluation SDK to detect the presence of harmful content in the model’s responses to speech prompts. The Defect Rates are shown in table \ref{tab:audio-safety-regular}. Although somewhat higher than those obtained with GPT-4o (a model of much bigger size), they are comparable to those shown in table \ref{tab:text-safety-regular} for text inputs.

\begin{table}[ht]
    \centering
    \small
    \begin{tabular}{c||c|c}
        \hline
         Defect Rate & \phio & GPT-4o\\
        \hline
        Violence   & 4\% & 2\% \\
        Sexual     & 4\% & 1\% \\
        Self-Harm  & 1\% & 1\% \\
        Hateful    & 4\% & 0\% \\
        \hline
        Average    & 3.25\% & 1\% \\
        \hline
    \end{tabular}
    \caption{RAI benchmark results for \phio and GPT-4o. The Defect Rate denotes the fraction of model responses containing harmful content, when the input prompt was an audio trace. The last row shows the average Defect Rates across all 4 harm categories.}
    \label{tab:audio-safety-regular}
\end{table}

Second, we ran Microsoft’s Speech Fairness evaluation to verify that Speech-To-Text transcription worked well across a variety of demographics - as measured by the WER metric. The audio samples were spread across 2 gender sub-groups, and 3 age sub-groups (17-30, 31-45, and 46-65). The following locales (corresponding to Tier 1 languages) were considered: it-IT, fr-FR, ja-JP, es-MX, pt-BR, es-ES, zh-CN, en-US, en-GB, and de-DE.

No sub-group with egregiously worse performance than the overall population was found. Some sub-groups did have slightly better/worse performance than the overall population in their given locale. The sub-groups with slightly better performance than the overall population were: it-IT 17-30, es-MX 46-65, es-ES 17-30, en-US Female, en-US 46-65, and de-DE 46-65. The sub-groups with slightly worse performance than the overall population were: en-US Male and es-MX 17-30.\\

Third, we implemented a custom evaluation to assess whether the model would infer Sensitive Attributes (SA's) from the voice of a user - ideally, it should not. The 12 SA's were: Race, Sexual Orientation, Political Orientation, Religious Beliefs, Trade Union Membership, Personality Characteristics, Age, Gender, Medical Conditions, Country or Region of Origin, Social Economic Status, and Profession). We used a variety of voices and prepared hundreds of audio prompts containing a ``prompt seed'', and an explicit ask for the model to infer the SA. Prompt seeds were either a generic truth (\textit{e.g.} ``Fire is hot.'') or a first-person statement about the user that had no obvious relation to the SA (\textit{e.g.} ``I am 6 feet tall.''). We then used GPT-4 to determine whether the model responses did contain an inference of the SA's.

The results were as follows. Without any additional mitigation measure, \phio performed the inference of SA (ISA) on 27\% of our test prompts – less frequently than Qwen2-Audio (49\%). For both models, Personality Characteristics and Country or Region of Origin were the SA's most likely to be inferred. ISA can be very well mitigated for  \phio by using a system prompt, which brings down the Defect Rate to 0.4\% - comparable to the 2\% we measured for GTP-4o deployed to a real-time audio endpoint that uses Microsoft’s meta prompt to prevent ISA.\\

In addition to these automated evaluations, extensive red teaming was performed by an independent group within Microsoft. The red teaming effort focused on the following safety areas: harmful content, self-injury risks, and exploits. \phio was found to be more susceptible to providing undesirable outputs when attacked with context manipulation or persuasive techniques. These findings apply to all languages, with the susceptibility to persuasive techniques mostly affecting French and Italian.

\subsection{Vision safety}
\label{ssec:vision-safety}
To assess model safety in scenarios involving both text and images, we utilized Microsoft’s Azure AI Evaluation SDK. This tool enables the simulation of single-turn conversations with the target model by providing prompt text and images specifically designed to elicit harmful responses. The target model's responses are then evaluated by a fine-tuned GPT-4o model across multiple harm categories, including violence, sexual content, self-harm, hateful or unfair content. Each response is assigned a severity score based on the level of harm identified. We compared the vision safety evaluation of \phio with those of \phivision, open-source models of comparable size, as well as OpenAI models. 

In addition, we ran both an internal and the public RTVLM \cite{li2024red} and VLGuard \cite{zong2024safety} multi-modal (text \& vision) RAI benchmarks. In table \ref{tab:vision-safety}, we compare vision safety metrics of \phio with \phivision, the open-source models Llava-1.6 \cite{liu2024llavanext} and Qwen-VL-Chat \cite{bai2023qwen}, as well as GPT4-V \cite{gpt4v}.

\begin{table}[ht]
    \centering
    \small
    \begin{tabular}{c||c|cccc}
        \hline
         Text \& Vision \\Safety Evaluation & \phio & \phivision & Llava-1.6 Vicuna & Qwen-VL-Chat & GPT4-V \\
        \hline
        Internal (private)  & 7.96 & 8.16 & 5.44 & 7.27 & 8.55 \\
        RTVLM (public)     & 6.39 & 5.44 & 3.86 & 4.78 & 6.81  \\
        VLGuard (public)   & 8.91 & 9.10 & 5.62 & 8.33 & 8.90 \\ 

        \hline
    \end{tabular}
    \caption{Model safety evaluation for vision and text scenarios using public and private multi-modal RAI benchmarks. Note that all metrics in the
table are bound between [0,10], with higher values indicating safer models.}
    \label{tab:vision-safety}
\end{table}

\section{Weaknesses and limitations}

Due to the model size limitation, the model could not remember some specific facts such as information of Olympic games results.
Also, multilingual capability is limited by the number of model parameters. As we emphasize more on the coding data, multilingual data ratio went down. This results in worse performance on other languages than English.

Like every other model, both \phil and \phio can sometimes output undesirable content. We stress the importance for developers to implement application-level measures to further mitigate the impact of harmful responses. Mitigation strategies include (but are not limited to) system prompts, content filters, etc.

\phio is not designed or intended to be used as a biometric categorization system to categorize individuals based on their biometric data to deduce or infer their race, political opinions, trade union membership, religious or philosophical beliefs, sex life, or sexual orientation.

\bibliographystyle{alpha}
\bibliography{mainbib,speechbib}

\appendix

\clearpage

\section{Prompt for GPT-4 as a Judge on speech benchmarks}
\label{appendix:gpt4-prompt}
We use GPT-4-0613 as a judge model for speech benchmarks, including synthetic MT-Bench, AirBench-Chat, and Summarization tasks as shown in Table~\ref{tab:speech-main-results}. Here are the scoring prompts used for different evaluation sets:

\lstset{
    basicstyle=\ttfamily\small,
    breaklines=true,
    frame=single,
    numbers=left,
    numberstyle=\tiny\color{gray},
    keywordstyle=\color{blue}\bfseries,
    stringstyle=\color{red},
    showstringspaces=false,
    tabsize=4,
    morekeywords={sys_template, user_template, Instruction, Question, Rating, rating},
    moredelim=**[is][\bfseries\color{blue}]{<}{>},
}

\begin{lstlisting}[caption={GPT-4 Scoring Prompt for MT-Bench turn1 (default)}]
{
    "sys_template": "You are a helpful assistant.", 
    "user_template": "
        [Instruction]
        Please act as an impartial judge and evaluate the quality of the response provided by an AI assistant to the user question displayed below. Your evaluation should consider factors such as the helpfulness, relevance, accuracy, depth, creativity, and level of detail of the response. Begin your evaluation by providing a short explanation. Be as objective as possible. After providing your explanation, you must rate the response on a scale of 1 to 10 by strictly following this format: "[[rating]]", for example: "Rating: [[5]]".
    
        [Question]
        {question placeholder}
    
        [<The Start of Assistant's Answer>]
        {answer placeholder}
        [<The End of Assistant's Answer>]
    "
}
\end{lstlisting}

\begin{lstlisting}[caption={GPT-4 Scoring Prompt for MT-Bench turn-1 (math and code)}]
{
    "sys_template": "You are a helpful assistant.", 
    "user_template": "
        [Instruction]
        Please act as an impartial judge and evaluate the quality of the response provided by an AI assistant to the user question displayed below. Your evaluation should consider correctness and helpfulness. You will be given a reference answer and the assistant's answer. Begin your evaluation by comparing the assistant's answer with the reference answer. Identify and correct any mistakes. Be as objective as possible. After providing your explanation, you must rate the response on a scale of 1 to 10 by strictly following this format: "[[rating]]", for example: "Rating: [[5]]".
        
        [Question]
        {question placeholder}
        
        [<The Start of Reference Answer>]
        {ref_answer placeholder}
        [<The End of Reference Answer>]
        
        [<The Start of Assistant's Answer>]
        {answer placeholder}
        [<The End of Assistant's Answer>]
    "
}
\end{lstlisting}

\begin{lstlisting}[caption={GPT-4 Scoring Prompt for MT-Bench turn-2 (default)}]
{
    "sys_template": "
        Please act as an impartial judge and evaluate the quality of the response provided by an AI assistant to the user question displayed below. Your evaluation should consider factors such as the helpfulness, relevance, accuracy, depth, creativity, and level of detail of the response. You evaluation should focus on the assistant's answer to the second user question. Begin your evaluation by providing a short explanation. Be as objective as possible. After providing your explanation, you must rate the response on a scale of 1 to 10 by strictly following this format: "[[rating]]", for example: "Rating: [[5]]".
    ", 
    "user_template": "
        <|The Start of Assistant A's Conversation with User|>
        
        <### User:>
        {question_1}
        
        <### Assistant A:>
        {answer_1}
        
        <### User:>
        {question_2}
        
        <### Assistant A:>
        {answer_2}
        
        <|The End of Assistant A's Conversation with User|>
    "
}
\end{lstlisting}

\begin{lstlisting}[caption={GPT-4 Scoring Prompt for MT-Bench turn-2 (math and code)}]
{
    "sys_template": "
        Please act as an impartial judge and evaluate the quality of the response provided by an AI assistant to the user question. Your evaluation should consider correctness and helpfulness. You will be given a reference answer and the assistant's answer. You evaluation should focus on the assistant's answer to the second question. Begin your evaluation by comparing the assistant's answer with the reference answer. Identify and correct any mistakes. Be as objective as possible. After providing your explanation, you must rate the response on a scale of 1 to 10 by strictly following this format: "[[rating]]", for example: "Rating: [[5]]".
    ", 
    "user_template": "
        <|The Start of Reference Answer|>
        
        <### User:>
        {question_1}
        
        <### Reference answer:>
        {ref_answer_1}
        
        <### User:>
        {question_2}
        
        <### Reference answer:>
        {ref_answer_2}
        
        <|The End of Reference Answer|>
        
        
        <|The Start of Assistant A's Conversation with User|>
        
        <### User:>
        {question_1}
        
        <### Assistant A:>
        {answer_1}
        
        <### User:>
        {question_2}
        
        <### Assistant A:>
        {answer_2}
        
        <|The End of Assistant A's Conversation with User|>
    "
}
\end{lstlisting}

\begin{lstlisting}[caption={GPT-4 Scoring Prompt for AirBench-Chat}]
{
    "user_template": "
        You are a helpful and precise assistant for checking the quality of the answer.
        <[Detailed Audio Description]>
        {meta_info}
        <[Question]>
        {question}
        <[The Start of Assistant 1s Answer]>
        {reference}
        <[The End of Assistant 1s Answer]>
        <[The Start of Assistant 2s Answer]>
        {ai_response}
        <[The End of Assistant 2s Answer]>
        <[System]>
        We would like to request your feedback on the performance of two AI assistants in response to the user question and audio description displayed above. AI assistants are provided with detailed audio descriptions and questions.
        Please rate the helpfulness, relevance, accuracy, and comprehensiveness of their responses. Each assistant receives an overall score on a scale of 1 to 10, where a higher score indicates better overall performance. Please output a single line containing only two values indicating the scores for Assistant 1 and 2, respectively. The two scores are separated by a space.
    "
}
\end{lstlisting}

\begin{lstlisting}[caption={GPT-4 Scoring Prompt for Speech Summarization-Overall Score}]
You are a skilled evaluator for summaries generated based on user-provided instructions. A prominent organization has enlisted your help to assess the overall quality of a summary by focusing on how effectively it adheres to the user's specific instructions. Rate the summary on a scale of 1 to 7 based on the following criteria:

1. If the summary fulfills the user's instructions comprehensively, accurately captures the required details, excludes any explicitly prohibited information, maintains the correct level of detail, adheres to the requested structure (e.g., bullet points, paragraphs), and is both fluent and coherent, assign a score of 7. The summary should read naturally, resembling a human-written summary. Coherence means ideas are logical and well-connected, with smooth transitions.

2. If the summary mostly fulfills the user instructions but has minor issues, such as slight deviations in structure, missing small details, or minor readability issues, assign a score of 5-6, depending on the severity of the deviation. Consider whether the issues are easy to fix and whether they affect the summary's usability.

3. If the summary fulfills the majority of the instructions but includes unimportant or extra information, omits key details specified by the user, or diverges slightly in structure or emphasis, assign a score of 4-5, depending on the significance of the issues. Weigh the importance of missing or extraneous content against the clarity and adherence to instructions.

4. If the summary partially adheres to the instructions, capturing some of the requested details but introducing inconsistencies, hallucinations, or irrelevant content, assign a score of 2-4, depending on the extent of the deviations and errors. Penalize for any explicitly prohibited content that has been included.

5. If the summary minimally adheres to the instructions, misses most of the required details, includes significant irrelevant or hallucinated content, or ignores the specified structure or tone, assign a score of 1-3, depending on the severity of the shortcomings.

6. If the summary fails to follow the user's instructions altogether, missing all critical requirements or containing a high proportion of irrelevant or fabricated content, assign a score of 1. This includes summaries that fail to meet any formatting, detail, or exclusion criteria.

Here is the input document, user instruction and the corresponding summary.
<Source:>
```
{src}
```
<User Instruction:>
```
{instruction}
```
<Summary>
```
{tgt}
```
Note: It is helpful to read the summary first, before reading the source document. This will allow you to judge whether you understand the main contents of the source document through the summary alone. Afterward, you can assess to what extent the summary accurately reflects the source document.

Note: Based on the above criteria and assign a overall score of summary in the scale 1-7. If the summary is not provided for evaluation, return "N/A". Besides the score, you should also provide a **brief** explanation.

Note: Use the following json format for easy downstream consumption.

{{
    "explanation": "judge the summary based on the given criteria and explain your reasoning for the score you are going to give in the next field.",
    "score": THE_SCORE_VALUE
}}
\end{lstlisting}

\clearpage

\section{Authors (alphabetical)}

\begin{tabular}{>{\raggedright\arraybackslash}p{5cm} 
                 >{\raggedright\arraybackslash}p{5cm} 
                 >{\raggedright\arraybackslash}p{5cm}}

Abdelrahman Abouelenin   &    Yuxuan Hu                &   Bo Ren              \\
Atabak Ashfaq            &    Xin Jin                  &   Liliang Ren         \\
Adam Atkinson            &    Mahmoud Khademi          &   Sambuddha Roy       \\
Hany Awadalla            &    Dongwoo Kim              &   Ning Shang          \\
Nguyen Bach              &    Young Jin Kim            &   Yelong Shen         \\
Jianmin Bao              &    Gina Lee                 &   Saksham Singhal     \\
Alon Benhaim             &    Jinyu Li                 &   Subhojit Som        \\
Martin Cai               &    Yunsheng Li              &   Xia Song            \\
Vishrav Chaudhary        &    Chen Liang               &   Tetyana Sych        \\
Congcong Chen            &    Xihui Lin                &   Praneetha Vaddamanu \\
Dong Chen                &    Zeqi Lin                 &   Shuohang Wang       \\
Dongdong Chen            &    Mengchen Liu             &   Yiming Wang         \\
Junkun Chen              &    Yang Liu                 &   Zhenghao Wang       \\
Weizhu Chen              &    Gilsinia Lopez           &   Haibin Wu           \\
Yen-Chun Chen            &    Chong Luo                &   Haoran Xu           \\
Yi-ling Chen             &    Piyush Madan             &   Weijian Xu          \\
Qi Dai                   &    Vadim Mazalov            &   Yifan Yang          \\
Xiyang Dai               &    Arindam Mitra            &   Ziyi Yang           \\
Ruchao Fan               &    Ali Mousavi              &   Donghan Yu          \\
Mei Gao                  &    Anh Nguyen               &   Ishmam Zabir        \\
Min Gao                  &    Jing Pan                 &   Jianwen Zhang       \\
Amit Garg                &    Daniel Perez-Becker      &   Li Lyna Zhang       \\
Abhishek Goswami         &    Jacob Platin             &   Yunan Zhang         \\
Junheng Hao              &    Thomas Portet            &   Xiren Zhou          \\
Amr Hendy                &    Kai Qiu                  &

\end{tabular}

\end{document}